\documentclass[
11pt,              % was 12 originally
letterpaper,       % legalpaper, a4paper, b5paper, ..., landscape
fleqn,
headnosepline,	   % headsepline line separating header from text body
oneside,           % twoside
onecolumn,         % twocolumn
openright,         % openany (new chapter w/any page rather than right)
noonelinecaption,
cleardoubleempty,
smallheadings]
{scrartcl}         % {scrbook}, {scrreprt}, {scrlttr2}
\addtokomafont{caption}{\sffamily\footnotesize}
\setkomafont{pagenumber}{\sffamily\footnotesize}
\addtokomafont{pagenumber}{\sffamily\footnotesize}
\setkomafont{captionlabel}{\sffamily\bfseries\footnotesize}
\setcapindent{0em} % no indentation

\usepackage{graphicx}
\usepackage{psfrag}
\usepackage{amsmath}
\usepackage{amssymb}
\usepackage{color}
\usepackage{xcolor}
\usepackage{rotate}
\usepackage{makeidx}
\usepackage{cite}
\usepackage{eufrak}
\usepackage[varumlaut]{yfonts}
\usepackage{helvet}   % Helvetica font as default sans serif
\usepackage{mathpazo}
\normalfont
\usepackage[T1]{fontenc}
\usepackage{textcomp}
\usepackage{longtable} 
\usepackage[only,llbracket,rrbracket]{stmaryrd}
\usepackage{bm}
\begin{document}
%%%%%%%%%%%%%%%%%%%%%%%%%%%%%%%%%%%%%%%%%%%%%%%%%%%%%%%%%%%%%%%%%%%%%%%%
\newcommand{\beq} {\begin{equation}}
\newcommand{\eeq} {\end{equation}}
\newcommand{\D}   {\displaystyle}
\newcommand{\divg}{\mbox{\rm{div}}\,}
\newcommand{\clearemptydoublepage}{\newpage{\pagestyle{empty}\cleardoublepage}}
\newcommand{\Divg}{\mbox{\rm{Div}}\,}
%%%%%%%%%%%%%%%%%%%%%%%%%%%%%%%%%%%%%%%%%%%%%%%%%%%%%%%%%%%%%%%%%%%%%%%%
\newtheorem{remark}      {\bf{\sffamily{Remark}}}
\newtheorem{definition}  {\bf{\sffamily{Definition}}}
%\numberwithin{remark}    {section}
%\numberwithin{definition}{chapter}
%\numberwithin{equation}  {section} 
%\renewcommand\figurename {\bf{\sffamily\footnotesize{Figure}}}
%\renewcommand\tablename  {\bf{\sffamily\footnotesize{Table}}}
\renewcommand{\sc}{}
\renewcommand{\Psi}{\psi}
\renewcommand{\varrho}{\vartheta}
\renewcommand{\arraystretch}{1.3}
\sloppy
%%%%%%%%%%%%%%%%%%%%%%%%%%%%%%%%%%%%%%%%%%%%%%%%%%%%%%%%%%%%%%%%%%%%%%%%
\def\sca   #1{\mbox{\rm #1}{}}
\def\mat   #1{\mbox{\bf #1}{}}
\def\vec   #1{\mbox{\boldmath $#1$}{}}
\def\ten   #1{\mbox{\boldmath $#1$}{}}
\def\scas  #1{\mbox{{\scriptsize{${\rm{#1}}$}}}{}}
\def\vecs  #1{\mbox{{\boldmath{\scriptsize{$#1$}}}}{}}
\def\tens  #1{\mbox{{\boldmath{\scriptsize{$#1$}}}}{}}
\def\up    #1{^{\mbox{\rm{\footnotesize{#1}}}}}
\def\down  #1{_{\mbox{\rm{\footnotesize{#1}}}}}
\def\ltr   #1{\mbox{\sf{#1}}}
\def\bltr  #1{\mbox{\sffamily{\bfseries{{#1}}}}}
%%%%%%%%%%%%%%%%%%%%%%%%%%%%%%%%%%%%%%%%%%%%%%%%%%%%%%%%%%%%%%%%%%%%%%%%%
\vspace*{1.0cm}
\begin{center}
{\sffamily\bfseries\Large{Bayesian Physics-Informed Neural Networks}}\\[4pt]
{\sffamily\bfseries\Large{for real-world nonlinear dynamical systems}}
\end{center}
%%%%%%%%%%%%%%%%%%%%%%%%%%%%%%%%%%%%%%%%%%%%%%%%%%%%%%%%%%%%%%%%%%%%%%%%%
%%%%%%%%%%%%%%%%%%%%%%%%%%%%%%%%%%%%%%%%%%%%%%%%%%%%%%%%%%%%%%%%%%%%%%%%
\vspace*{0.1cm}
\begin{center}
Kevin Linka$^1$,
Amelie Sch\"afer$^2$,
Xuhui Meng$^3$,\\
Zongren Zou$^3$,
George Em Karniadakis$^3$,
Ellen Kuhl$^2$ 
\end{center}
%\vspace*{0.5cm}
\begin{center}
{\small{
$^1$ Institute of Continuum and Material Mechanics, \\
Hamburg University of Technology, Hamburg, Germany \\
$^2$ Department of Mechanical Engineering, \\
Stanford University, Stanford, California, United States \\
$^3$ Division of Applied Mathematics, \\
Brown University, Providence, Rhode Island, United States
}}
\end{center}
\vspace*{0.35cm}
%%%%%%%%%%%%%%%%%%%%%%%%%%%%%%%%%%%%%%%%%%%%%%%%%%%%%%%%%%%%%%%%%%%%%%%%
{\sffamily{\bfseries{Abstract.}}}
%%%%%%%%%%%%%%%%%%%%%%%%%%%%%%%%%%%%%%%%%%%%%%%%%%%%%%%%%%%%%%%%%%%%%%%%
Understanding real-world dynamical phenomena 
remains a challenging task.
Across various scientific disciplines, 
machine learning has advanced as
% machine learning has advanced as 
% a powerful technology 
% an indispensible
the go-to technology to 
analyze nonlinear dynamical systems,
identify patterns in big data, and 
make decision around them. 
%
%Traditional machine learning 
%offers powerful tools 
%to fit and describe real-world observations.
%
Neural networks % in particular 
are now consistently used as
universal function approximators for data with underlying mechanisms
that are incompletely understood or exceedingly complex.
However, neural networks alone
ignore the fundamental laws of physics 
and often fail to make plausible predictions.  
Here we integrate data, physics, and uncertainties
by combining neural networks, physics-informed modeling, and Bayesian inference
to improve the predictive potential 
of traditional neural network models.
We embed the physical model of a damped harmonic oscillator
into a fully-connected feed-forward neural network
to explore a simple and illustrative model system, 
the outbreak dynamics of COVID-19.
Our Physics-Informed Neural Networks can
seamlessly integrate data and physics, 
robustly solve forward and inverse problems, and 
perform well for both interpolation and extrapolation,
even for a small amount of noisy and incomplete data.
At only minor additional cost, 
they can self-adaptively learn the weighting between data and physics.
% and perform well, even in regions with steep gradients.  
%
Combined with Bayesian Neural Networks, 
they can serve as priors in a Bayesian Inference, and
provide credible intervals for uncertainty quantification. 
Our study reveals the inherent advantages and disadvantages 
of Neural Networks, Bayesian Inference, and a combination of both 
and provides valuable guidelines for model selection. 
While we have only demonstrated these different approaches 
for the simple model problem of a seasonal endemic infectious disease, 
we anticipate that the underlying concepts and trends generalize to more complex disease conditions and, more broadly, to a wide variety of nonlinear dynamical systems. 

\vspace*{0.5cm}

\clearpage
%%%%%%%%%%%%%%%%%%%%%%%%%%%%%%%%%%%%%%%%%%%%%%%%%%%%%%%%%%%%%%%%%%%
%\linenumbers
\section{Motivation}\label{motiv}
%%%%%%%%%%%%%%%%%%%%%%%%%%%%%%%%%%%%%%%%%%%%%%%%%%%%%%%%%%%%%%%%%%%
% machine learning > physics
%%%%%%%%%%%%%%%%%%%%%%%%%%%%%%%%%%%%%%%%%%%%%%%%%%%%%%%%%%%%%%%%%%%
\noindent
Modeling and predicting the behavior of complex nonlinear dynamical systems remains a challenging scientific task \cite{peng21}. 
Numerous scientists are addressing this challenge by collecting more observational data than ever before; however, more often than not without a clear picture how to sort, analyze, and understand these enormous amounts of information \cite{karniadakis21}. 
Throughout the past two decades, machine learning has advanced as the go-to technology to analyze big data and explore the massive design spaces associated with them \cite{alber19}.
Simply put, machine learning is an incredibly powerful strategy to make data-driven recommendations and decisions based on the input data alone.
As such, it is has become an indispensable technology for image or speech recognition, medical diagnostics, or self-driving cars.  
However, traditional machine learning ignores the fundamental laws of physics, and, as a consequence, performs well at fitting observations, but often fails to make consistent and plausible predictions.  
In the engineering science community, especially in disciplines that are traditionally not data-rich, these limitations have raised the question how we can advance machine learning and build in our prior knowledge to respect underlying physical principles.\\[6.pt]
%%%%%%%%%%%%%%%%%%%%%%%%%%%%%%%%%%%%%%%%%%%%%%%%%%%%%%%%%%%%%%%%%%%
% (physics-informed) neural networks
%%%%%%%%%%%%%%%%%%%%%%%%%%%%%%%%%%%%%%%%%%%%%%%%%%%%%%%%%%%%%%%%%%%
%* classical computational methods fail to seamlessly incorporate noise data into   
%  existing algorithms
%* machine learning is emerging as a promising strategy
%* but training deep neural networks requires big data
%* rather than relying on big data, 
%  we can integrate additional constraints by enforcing physical laws
The trend of embedding physics into machine learning has recently drawn tremendous attention in various scientific applications \cite{raissi17}.
As engineering scientists, we have been modeling complex nonlinear dynamical systems for hundreds of years and we have developed a reasonable quantitative knowledge about physical parameters, boundary conditions, and constraints \cite{raissi18}. 
Physics-informed machine learning now allows us to use this prior knowledge when training machine learning tools \cite{alber19}. 
This not only makes the training more efficient, but also more accurate and robust, especially when working with missing, noisy, or sparse real-life data \cite{yang19}. 
A powerful and effective strategy to seamlessly integrate data and physics are Physics-Informed Neural Networks \cite{raissi19}. 
% neural network-based regression offers effective, simple, and meshless 
% framework to implement physics informed learning machines
Physics Informed Neural Networks introduce a learning bias by directly embedding the physics into the loss function of a neural network \cite{raissi21}.
%- appropriate choice of loss function to modulate the training phase and explicitly 
% favor convergence towards a solution that adheres to the underlying physics
%- calculate gradients via automatic differentiation.
%- are effective and efficient for ill-posed and inverse problems
This makes the neural network more robust, especially in the presence of sparse or imperfect data, and can provide more accurate and physically consistent predictions, even outside the training window    
\cite{karniadakis21}.
For example, recent studies have successfully applied Physics Informed Neural Networks to study the complex outbreak dynamics of COVID-19 \cite{cai22,treibert21}
by integrating advance epidemiology models into deep neural networks \cite{kharazmi21,zhang21}.
However, the success of these methods depends crucially on the amount of available data and the complexity of the system itself. 
For the COVID-19 pandemic, we now know that predictions typically fail beyond a two-week window \cite{bhouri21,linka21}, and that predicting outbreak dynamics beyond this range can have devastating consequences \cite{holmdahl20}. 
It seems intuitive to ask how we can quantify uncertainties to estimate the reliability of our models and to build confidence in our model predictions  \cite{meng21}.\\[6.pt]
%%%%%%%%%%%%%%%%%%%%%%%%%%%%%%%%%%%%%%%%%%%%%%%%%%%%%%%%%%%%%%%%%%%
% bayesian inference
%%%%%%%%%%%%%%%%%%%%%%%%%%%%%%%%%%%%%%%%%%%%%%%%%%%%%%%%%%%%%%%%%%%
Real-world data are inherently stochastic, noisy, and incomplete, and naturally contain aleatoric uncertainty \cite{oden17}. 
This is why we should never just look at the plain data.
Instead, we should always interpret data in the context of models that allow us to  quantifying uncertainties \cite{oden10}.
A recently proposed promising strategy is to interpret Physics Informed Neural Networks within the framework of Bayesian statistics \cite{yang19}. 
This approach combines a Bayesian Neural Network with a Physics Informed Neural Network to create prior distributions and uses Hamilton Monte Carlo methods to estimate posterior distributions \cite{yang21}.  
Uncertainty quantification in the form of credible intervals is a natural and important by-product of such a Bayesian analysis \cite{gelman13}.
Throughout the past decades, Bayesian statistics have undeniably gained massive popularity \cite{oden18} and have stimulated the birth to an entirely new field, probabilistic programming \cite{osvaldo18}. 
Combined with Physics Informed Neural Networks, Bayesian statistics perform well for both forward and inverse problems including Poisson problems, flow problems, phase separation problems, and reaction-diffusion problems \cite{yang21}. 
However, most of these applications train their network on synthetic data from a known solution with overlaid noise \cite{meng22}.
This motivates the question how well Bayesian Physics Informed Neural Networks perform on real-word data from a nonlinear dynamical system for which the underlying physics are not entirely known, with both aleatoric and epistemic uncertainty.\\[6.pt]
%%%%%%%%%%%%%%%%%%%%%%%%%%%%%%%%%%%%%%%%%%%%%%%%%%%%%%%%%%%%%%%%%%%
% bayesian inference
%%%%%%%%%%%%%%%%%%%%%%%%%%%%%%%%%%%%%%%%%%%%%%%%%%%%%%%%%%%%%%%%%%%
The objective of our study is to illustrate the performance of Neural Networks, Bayesian Inference, and a combination of both using real-world data.
However, rather than showcasing these methods on real big data, we focus on a simple data set that is easy to understand, interpret, and reproduce:
the reported daily number of new COVID-19 cases worldwide \cite{linka20a} throughout the year of 2021 \cite{johnshopkins21}.
We identify a suitable, yet simple mathematical model with an analytical solution that allows us to characterize the data.
Again, our objective is not to develop the most rigorous physical model to explain each and every feature of the outbreak dynamics of a global pandemic \cite{jha20}. 
Instead, we adopt simple physical model, a damped harmonic oscillator, to characterize the long-term dynamics of COVID-19 on a global scale.
We integrate the case data and the oscillator model using Neural Networks and Bayesian Inference and compare their model equations, parameters, training, predictions, and uncertainties. 
We envision that this study sheds light on the performance of different cutting-edge machine learning tools, highlights their advantages and disadvantages, and helps engineering scientists to select the best method to analyze real-world nonlinear dynamical systems.
%%%%%%%%%%%%%%%%%%%%%%%%%%%%%%%%%%%%%%%%%%%%%%%%%%%%%%%%%%%%%%%%%%%
\section{Model problem}\label{model}
%%%%%%%%%%%%%%%%%%%%%%%%%%%%%%%%%%%%%%%%%%%%%%%%%%%%%%%%%%%%%%%%%%%
%%%%%%%%%%%%%%%%%%%%%%%%%%%%%%%%%%%%%%%%%%%%%%%%%%%%%%%%%%%%%%%%%%%%%%%%
\subsection{Data - Daily new COVID-19 cases}  
\begin{figure}
\centering
\includegraphics[width=0.5\linewidth]{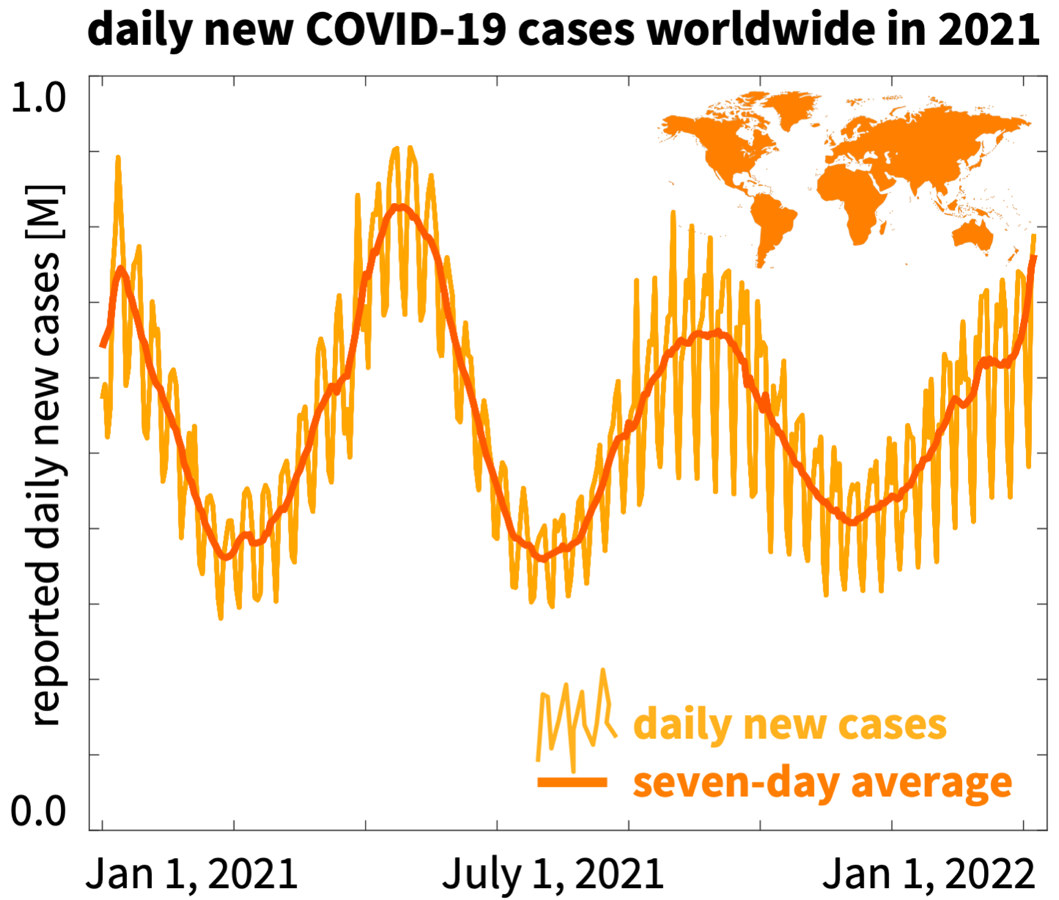}
\caption{{\bf{\sffamily{Data - Daily new COVID-19 cases worldwide in 2021.}}} 
Thin fluctuating line illustrates the reported daily new cases of COVID-19 worldwide;
thick smooth line illustrates the seven-day moving average of the daily new cases,
from January 1, 2021 to December 31, 2021.}
\label{fig01}
\end{figure}

% https://covariants.org/per-country
%%%%%%%%%%%%%%%%%%%%%%%%%%%%%%%%%%%%%%%%%%%%%%%%%%%%%%%%%%%%%%%%%%%%%%%%
\noindent
Figure \ref{fig01} illustrates our real-world data to compare different machine learning approaches. As example, we use the number of new COVID-19 cases worldwide \cite{peirlinck20} throughout the entire year of 2021, which we draw from a public database \cite{johnshopkins21}. 
The thin fluctuating line shows the raw data, the reported daily new cases worldwide
from January 1, 2021 to December 31, 2021. 
Global reporting started on January 22, 2020 with a total number of cases of 557 \cite{linka20}. 
On January 1, 2021, at the beginning of our analysis window, 
the number of newly reported cases worldwide was 572,602 bringing the total number of cases up to 84,332,767. 
On December 31, 2021, at the end of our window, the total number of cases was 288,631,129.
Altogether, the reporting window saw three pronounced waves. 
The first wave had 
a minimum number of new cases of 281,223 on day 46, February 16, 2021, and
a maximum number of 905,378 on day 118, April 29, 2021, 
as we can confirm from the global minimum and maximum in Figure \ref{fig01}.
The second wave had a 
a minimum number of new cases of 296,808 on day 172, June 22, 2021, and
a maximum number of 819,336 cases on day 221, August 10, 2021,
about four months after. 
To eliminate noise, reporting uncertainties, and weekday-weekend fluctuations, data analysts typically average the reported case numbers across a seven-day window \cite{kuhl21}. The thick smooth line illustrates the seven-day moving average $\hat{x}(t)$ of the daily new cases. 
In the following sections, we will use both the raw data and their seven-day moving average as an example to illustrate the potential of Neural Networks, Bayesian Inference, and both methods combined. 
%%%%%%%%%%%%%%%%%%%%%%%%%%%%%%%%%%%%%%%%%%%%%%%%%%%%%%%%%%%%%%%%%%%%%%%%
%%%%%%%%%%%%%%%%%%%%%%%%%%%%%%%%%%%%%%%%%%%%%%%%%%%%%%%%%%%%%%%%%%%%%%%%
%%%%%%%%%%%%%%%%%%%%%%%%%%%%%%%%%%%%%%%%%%%%%%%%%%%%%%%%%%%%%%%%%%%%%%%%
\subsection{Neural Network modeling - Fully-connected feed-forward neural network}\label{neural_networks}
%%%%%%%%%%%%%%%%%%%%%%%%%%%%%%%%%%%%%%%%%%%%%%%%%%%%%%%%%%%%%%%%%%%%%%%%
\noindent
To model the data in Figure \ref{fig01} 
without any prior knowledge of the underlying physics, 
we use a neural network 
that approximates the model solution $\mat{x}$,
the reported daily new COVID-19 cases,
by taking the time coordinates $\mat{t}$ as input.
We adopt one of the simplest examples of neural networks, 
a fully-connected feed-forward neural network
composed of multiple hidden layers.
We denote the hidden variables of the $k^{\rm{th}}$ hidden layer as $\mat{z}_{k}$
and summarize the neural network, here for two hidden layers, as 
\beq
\begin{array}{l@{\hspace*{0.2cm}}c@{\hspace*{0.2cm}}
              l@{\hspace*{0.2cm}}l@{\hspace*{0.2cm}}
              l@{\hspace*{0.2cm}}l}
\mat{z}_0 &=&            &                       &  &\mat{t} \\
\mat{z}_1 &=& \sigma \,( &\mat{W}_1 \, \mat{z}_0 &+ &\mat{b}_1\,)  \\
\mat{z}_2 &=& \sigma \,( &\mat{W}_2 \, \mat{z}_1 &+ &\mat{b}_2\,)  \\
\mat{z}_3 &=&            &\mat{W}_3 \, \mat{z}_2 &+ &\mat{b}_3  
\end{array}
\label{neuralnetwork}
\eeq
where the output of the last layer is used to approximate the true solution,
$\mat{x} \approx \mat{z}_3$.
Here,
$\sigma \,(\circ)$ denotes the non-linear activation function for which we select a hyperbolic tangent function, $\sigma \,(\circ)= \tanh\,(\circ)$,
and $\vec{\theta} = \{ \mat{W}_k, \mat{b}_k \}$
are the trainable network parameters,
the weight matrix $\mat{W}_k$ 
and the bias vector $\mat{b}_k$ of the $k^{th}$ layer of the network.
Throughout this project, 
for illustrative purposes,
we consider a fully-connected feed-forward network 
with one input, the time $t$, two hidden layers with 32 nodes each, and one output, the daily new cases $x(t)$. 
For this type of network,
$\mat{W}_1 \in \mathbb{R}^{1 \times 32}$,
$\mat{W}_2 \in \mathbb{R}^{32 \times 32}$,
$\mat{W}_3 \in \mathbb{R}^{32 \times 1}$,
and 
$\mat{b}_1 \in \mathbb{R}^{32}$,
$\mat{b}_2 \in \mathbb{R}^{32}$,
$\mat{b}_3 \in \mathbb{R}^{1}$,
resulting in
$32+32\times32+32 = 1088$ weights and $32+32+1 =65$ biases. 
%%%%%%%%%%%%%%%%%%%%%%%%%%%%%%%%%%%%%%%%%%%%%%%%%%%%%%%%%%%%%%%%%%%%%%%%
\subsection{Physic Informed modeling - Damped harmonic oscillator}\label{physics}
%%%%%%%%%%%%%%%%%%%%%%%%%%%%%%%%%%%%%%%%%%%%%%%%%%%%%%%%%%%%%%%%%%%%%%%%
\noindent 
To model the data in Figure \ref{fig01} using a physics-based model, 
we consider a general parameterized second order differential equation,
\beq
r(t,x,\dot{x},\ddot{x},\vec{\vartheta}) = 0  
\label{residual}
\eeq
where 
$t\in [\,0,T\,]$ is the time, 
$x(t)$ is the solution  
with initial conditions $x(t_0) = x_0$ and $\dot{x}(t_0) = \dot{x}_0$,
$r$ is the residual as a function of $x$ and its first and second time derivatives $\dot{x}$ and $\ddot{x}$, and
$\vec{\vartheta}$ are the physics parameters.
We now specify the general equation (\ref{residual}) for the model problem of a damped harmonic oscillator, which is governed by Newton's second law, the balance of forces,  
$ m \, \ddot{x} 
+ c \, \dot{x} 
+ k \,  x = 0 $,
or, divided by the mass $m$,
\beq
  r
= \ddot{x} 
+ \frac{c}{m} \, \dot{x} 
+ \frac{k}{m} \,  x = 0  
  \qquad\mbox{or}\qquad
  r
= \ddot{x} 
+ 2\, \zeta \omega_0 \, \dot{x} 
+ \omega_0^2 \, x  = 0 \,.
\label{oscillator}
\eeq
We can parameterize the damped harmonic oscillator equation in terms of 
the mass $m$,
the viscous damping coefficient $c$, and
the stiffness $k$, or, equivalently, in terms of 
the damping ratio $\zeta = \delta/\omega_0=c/(2\sqrt{mk})$, 
the damping $\delta=c/(2m)$, and
the angular frequency $\omega_0=\sqrt{k/m}$.
Equations (\ref{oscillator}.1) and (\ref{oscillator}.2) are homogeneous linear differential equations of second order with constant coefficients. For solutions to this type of equations, we can make an exponential ansatz of the form 
$       x(t) = C\, \exp(\lambda\,t)$, such that
$ \dot{x}(t) = \lambda \, C\, \exp(\lambda\,t)$ and
$\ddot{x}(t) = \lambda^2 C\, \exp(\lambda\,t)$.
Inserting this ansatz into equations (\ref{oscillator}.1) and (\ref{oscillator}.2) yields their characteristic equations,
\beq
  \lambda^2
+ \frac{c}{m} \, \lambda 
+ \frac{k}{m} = 0 
  \qquad \mbox{or} \qquad
  \lambda^2
+ 2\,\delta \, \lambda 
+ \omega_0^2 = 0 \,.
\label{oscillator03}
\eeq
These equations are quadratic equations in normal form with two solutions for $\lambda$,
\beq
\lambda_{1,2} = -\frac{c}{2m} \pm \sqrt{\left(\frac{c}{2m}\right)^2-\frac{k}{m}} 
\qquad \mbox{or} \qquad
\lambda_{1,2} = -\delta \pm \sqrt{\delta^2-\omega_0^2} \,.
\label{lambda}
\eeq
From equation (\ref{lambda}.2) we conclude that  
the damping ratio $\zeta= \delta/\omega_0$,
the ratio between damping $\delta$ and frequency $\omega_0$,
determines the type of the solution and with it the behavior of the system:
A damped harmonic oscillator can be 
overdamped for        $\zeta > 1$ and $\delta > \omega_0$ 
with two different real valued solutions for $\lambda$; 
critically damped for $\zeta = 1$ and $\delta = \omega_0$ 
with two identical real valued solutions for $\lambda$, or 
underdamped for       $\zeta < 1$ and $\delta < \omega_0$ 
with two conjugate complex solutions for $\lambda$. 
Here we consider the underdamped case. 
The analytical solution of an underdamped harmonic oscillator describes an exponential decay of the oscillation,
\beq
x(t) = 2A_0 \, \cos(\omega\,t + \phi) \,\exp \,(-\delta\,t) \,,
\label{analytical}
\eeq
where 
$A_0$ is the amplitude,
$\omega =\sqrt{\omega_0^2 - \delta^2}$ is the natural frequency, and 
$\phi= \arctan \,(-{\delta}/{\omega})$ is the phase angle.
%%%%%%%%%%%%%%%%%%%%%%%%%%%%%%%%%%%%%%%%%%%%%%%%%%%%%%%%%%%%%%%%%%%%%%%%
\begin{figure}[t]
\centering
\includegraphics[width=0.5\linewidth]{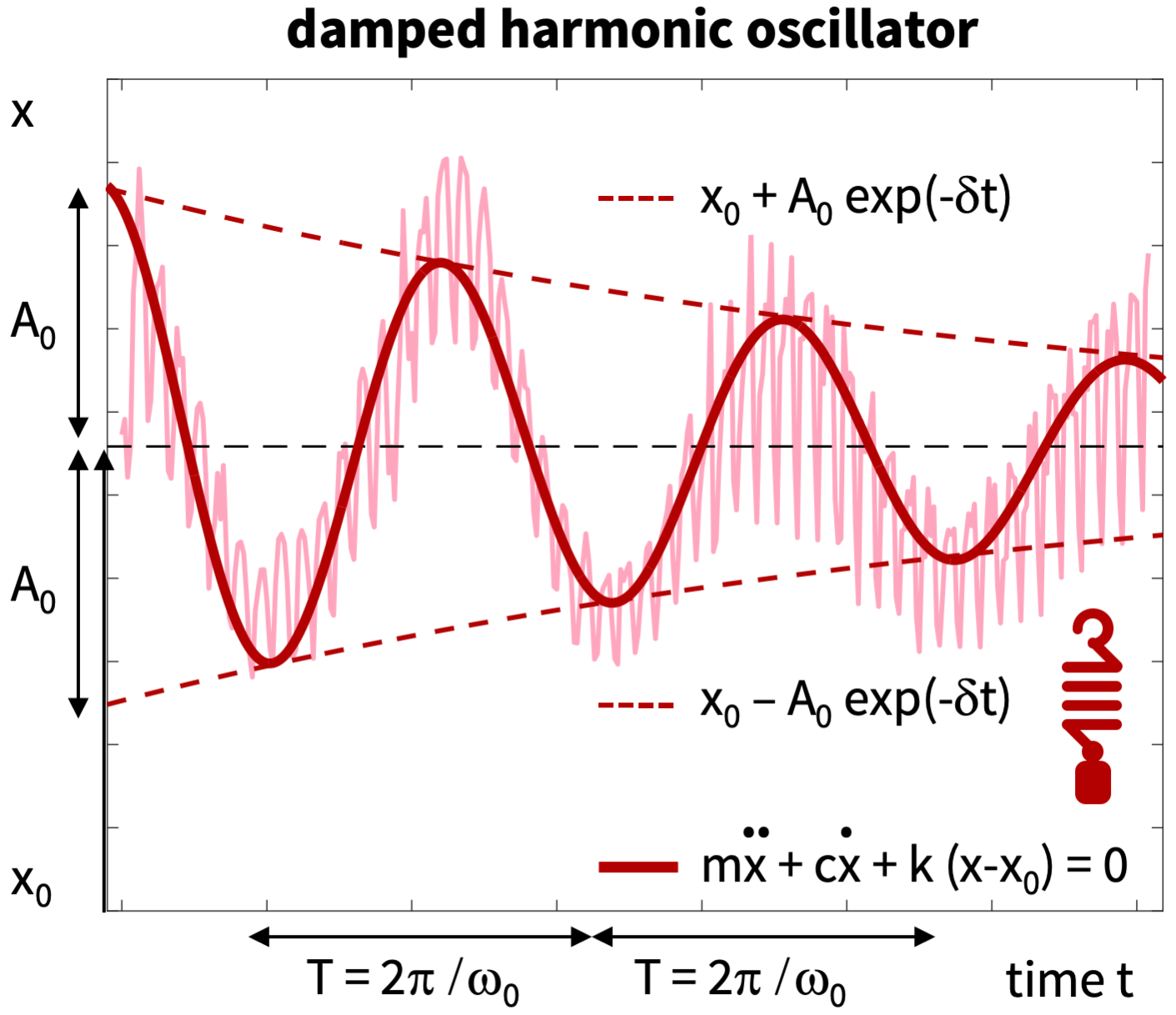}
\caption{{\bf{\sffamily{Physics - Damped harmonic oscillator.}}} 
Characteristic exponential decay of the oscillation 
in terms of the
period $T = 2 \pi / \omega_0$ with frequency $\omega_0 = \sqrt{k/m}$,
damping $\delta = c/(2m)$,
offset $x_0$, and
amplitude $A_0$.
Solid lines illustrate the analytical solution
$m\, \ddot{x} + c \, \dot{x} + k (x - x_0) = 0$ or
$\ddot{x} + 2\, \zeta \omega_0 \, \dot{x} + \omega_0^2 \, (x - x_0) = 0$, 
dashed lines illustrate the exponential decay in amplitude 
$x_0 \pm A_0 \, \exp(-\delta t)$ around the offset $x_0$.}
\label{fig02}
\end{figure}\\[4.pt]
%%%%%%%%%%%%%%%%%%%%%%%%%%%%%%%%%%%%%%%%%%%%%%%%%%%%%%%%%%%%%%%%%%%%%%%%
\noindent 
Figure \ref{fig02} illustrates the dynamics of a damped harmonic oscillator. 
The solid line highlights the analytical solution
$m\, \ddot{x} + c \, \dot{x} + k (x - x_0) = 0$ or
$\ddot{x} + 2\, \zeta \omega_0 \, \dot{x} + \omega_0^2 \, (x - x_0) = 0$, 
the dashed lines highlight the exponential decay in amplitude, 
$x_0 \pm A_0 \, \exp(-\delta t)$, where
$A_0$ is the amplitude,
$\delta = c/(2m)$ is the damping, 
$\omega_0 = \sqrt{k/m}$ is the frequency associated with the period
$T = 2 \pi / \omega_0$, and
$x_0$ is the offset.
We fix the value of the mass to $m=1$ and collectively summarize the remaining physics parameters in the parameter vector 
$\vec{\vartheta} = \left\{ c,k,x_0 \right\}$.
%$\vec{\vartheta} = \left\{ A_0, c,k,x_0 \right\}$.
%%%%%%%%%%%%%%%%%%%%%%%%%%%%%%%%%%%%%%%%%%%%%%%%%%%%%%%%%%%%%%%%%%%%%%%%
\subsection{Machine Learning - Integrating data and physics} 
%%%%%%%%%%%%%%%%%%%%%%%%%%%%%%%%%%%%%%%%%%%%%%%%%%%%%%%%%%%%%%%%%%%%%%%%
\noindent
To create learning machines that seamlessly integrate
the data from Figure \ref{fig01}, 
the physics from Figure \ref{fig02}, or both,
we combine 
Neural Network modeling from Section \ref{neural_networks} and 
Physics Informed modeling from Section \ref{physics}
and learn the underlying network and physics parameters. 
To quantify the uncertainty of the learned parameters, 
and with it the quality of the model, 
we can integrate each of these methods into a Bayesian Inference. 
%%%%%%%%%%%%%%%%%%%%%%%%%%%%%%%%%%%%%%%%%%%%%%%%%%%%%%%%%%%%%%%%%%%
\begin{figure}[h]
\centering
\includegraphics[width=0.8\linewidth]{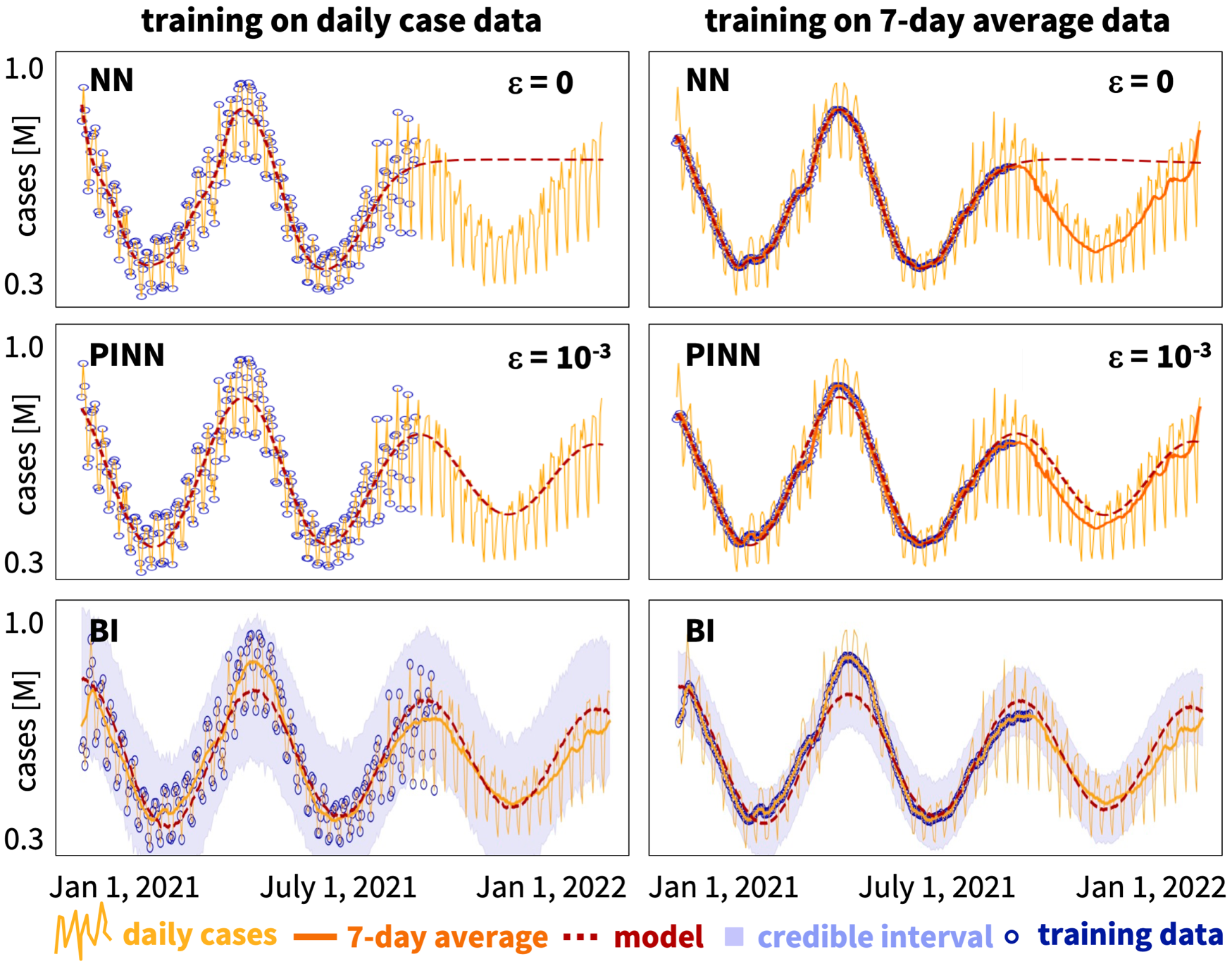}
\caption{{\bf{\sffamily{Machine learning - Integrating data and physics.}}}
Neural Networks approximate the training data well, but fail to predict the behavior outside the training window, when trained on both, the daily case data (top left) and the seven-day average (top right).
Physics Informed Neural Networks approximate both the training data and the behavior outside the training window, when trained on both, the daily case data (middle left) and the seven-day average data (middle right).
Bayesian Inference not only approximates both, the training data and the behavior outside the training window, but also provides credible intervals, which are wider when model parameters are inferred using the daily case data (bottom left) than when using the seven-day average data (bottom right).
Thin yellow lines indicate daily new case data, thick orange lines are their seven-day average, dashed red lines are the model simulations, light blue areas are its credible intervals, blue dots are the training data, here for 225 out of 365 days.} 
\label{fig03}
\end{figure}\\[6.pt]
%%%%%%%%%%%%%%%%%%%%%%%%%%%%%%%%%%%%%%%%%%%%%%%%%%%%%%%%%%%%%%%%%%%
Figure \ref{fig03} 
motivates the use of classical Neural Networks, Physics Informed Neural Networks, and classical Bayesian Inference before we describe these methods in detail in Sections \ref{NNmodeling} and \ref{BImodeling}.
The thin yellow lines indicate daily new case data, the thick orange lines are their seven-day average, the blue dots are the training data, here for 225 out of 365 days,
the dashed red lines are the model simulations, and light blue areas are the credible intervals of the model prediction. 
The left column uses the thin yellow lines of the daily new case data, and
the right column uses the thick orange lines of their seven-day average for training the models. \\[6.pt]
%%%
Neural Networks in the top row approximate the training data of the first 225 days well, but fail to predict the behavior outside the training window of the remaining 140 days. The neural network model simply continues the learned trend. Since day 225 corresponds to a plateau with a horizontal tangent, the neural network model predicts a flatline with constant case numbers for the remaining part of the year. This trend is similar when trained on both the daily case data in the top left and the seven-day average in the top right.\\[6.pt]
%%%
Physics Informed Neural Networks in the middle row approximate both the training data of the first 225 days and the behavior outside the training window of the remaining 140 days with a good accuracy. Throughout the entire year, the red dashed lines of the model remain close to the orange line of the seven-day case average. The fit of the Physics Informed Neural Network is virtually identical for training on the daily case data in the middle left and on the seven-day average in the middle right.\\[6.pt]
%%%
Bayesian Inference in the bottom row not only approximates both, the training data and the behavior outside the training window, but also provides credible intervals to estimate the quality of the fit. Since the Bayesian Inference uses the same underlying physics as the Physics Informed Neural Network, its red dashed lines also remain close to the orange line of the seven-day case average throughout the entire year. However, the fit with respect to the training data is worse compared to the Physics Informed Neural Network, because we bias the solution towards the physics equations. In contrast to the previous two models in the top and middle rows, the Bayesian Inference displays a clear difference between training on raw versus averaged data. The light blue areas of its credible intervals are notably wider when the model parameters are inferred using the daily case data in the bottom left than when using the seven-day average data in the bottom right.
%%%
We will now describe details of the underlying equations for Neural Networks in Section \ref{NNmodeling} and for Bayesian Inference in Section \ref{BImodeling}.
%%%%%%%%%%%%%%%%%%%%%%%%%%%%%%%%%%%%%%%%%%%%%%%%%%%%%%%%%%%%%%%%%%%
\section{Neural Network modeling}\label{NNmodeling}
%%%%%%%%%%%%%%%%%%%%%%%%%%%%%%%%%%%%%%%%%%%%%%%%%%%%%%%%%%%%%%%%%%%
%%%%%%%%%%%%%%%%%%%%%%%%%%%%%%%%%%%%%%%%%%%%%%%%%%%%%%%%%%%%%%%%%%%
\noindent The objective of Neural Network modeling is to train the neural network (\ref{neuralnetwork}) such that the model output $\mat{x}(\mat{t})$ best approximates the data $\hat{\mat{x}}$, by minimizing a loss function,
\beq
%L (\vec{\theta},\vec{\vartheta},\varepsilon \,;t) 
L (\vec{\Theta} \,;t) 
\rightarrow \mbox{min} \,,
\label{loss_general}
\eeq
through iteratively updating
the set of model parameters $\vec{\Theta}$.
The set of model parameters
could consist of 
the network parameters $\vec{\theta}$,
the physics parameters $\vec{\vartheta}$,
and a weighting coefficient $\varepsilon$,
\beq
\vec{\Theta} = \{ \vec{\theta}, \vec{\vartheta}, \varepsilon \}
\qquad \mbox{with} \qquad
\vec{\theta}= \{ \mat{W}_k, \mat{b}_k \} 
\qquad \mbox{and} \qquad
\vec{\vartheta}= \{ c, k, x_0 \},
%\vec{\vartheta}= \{ A_0, c, k, x_0 \},
\label{parameters_general}
\eeq
where
$\mat{W}_k$ and $\mat{b}_k$ are the network weights and biases 
from equation (\ref{neuralnetwork}) and
% $A_0$,
$c$, $k$, and $x_0$ are the physical damping, stiffness, and offset
from equation (\ref{oscillator}). 
We specify the loss function (\ref{loss_general}) and the relevant parameters (\ref{parameters_general})
for classical Neural Networks in Section \ref{NN},
for Physics Informed Neural Networks in Section \ref{PINN}, and
for Self Adaptive Physics Informed Neural Networks in Section \ref{SAPINN}.
%%%%%%%%%%%%%%%%%%%%%%%%%%%%%%%%%%%%%%%%%%%%%%%%%%%%%%%%%%%%%%%%%%%
\subsection{Neural Networks}\label{NN}
%%%%%%%%%%%%%%%%%%%%%%%%%%%%%%%%%%%%%%%%%%%%%%%%%%%%%%%%%%%%%%%%%%%
\noindent 
The objective of classical Neural Networks is to learn the network parameters $\vec{\theta}$, without prior knowledge of the underlying physics, by minimizing a loss function (\ref{loss_general}) that consists of a single term $L_{\rm{data}}$,
\beq
  L (\vec{\theta}\,; t) 
= L_{\rm{data}} %(\vec{\theta};t) 
\rightarrow \mbox{min} \,.
\label{loss_NN}
\eeq
The data loss $L_{\rm{data}}$ penalizes the error between model $\mat{x}(\mat{t})$ and data $\hat{\mat{x}}$.
We define it as the mean square error, %MSE, 
the $L_2$-norm of the difference between model and data,
$||\, \mat{x}(\mat{t}) - \hat{\mat{x}}\,||^2$, 
divided by the number of training points $n_{\rm{trn}}$,
\beq
  L_{\rm{data}} (\vec{\theta}\,; t)
= \frac{1}{n_{\rm{trn}}} \sum_{i=1}^{n_{\rm{trn}}}
|| \, x(t_i) - \hat{x}_i \, ||^2 \,.
\label{loss_data_NN}
\eeq
We minimize the loss function (\ref{loss_NN}) in terms of the mean square error (\ref{loss_data_NN}) to optimize the network parameters $\vec{\theta}$, 
the network weights and biases $\mat{W}_k$ and $\mat{b}_k$,
\beq
\vec{\theta} = \{ \mat{W}_k, \mat{b}_k \}\, ,
\eeq
using the ADAM optimizer, 
an adaptive algorithm for gradient-based first-order optimization. 
%%%%%%%%%%%%%%%%%%%%%%%%%%%%%%%%%%%%%%%%%%%%%%%%%%%%%%%%%%%%%%%%%%%
\subsection{Physics Informed Neural Networks}\label{PINN}
%%%%%%%%%%%%%%%%%%%%%%%%%%%%%%%%%%%%%%%%%%%%%%%%%%%%%%%%%%%%%%%%%%%
\noindent 
The objective of Physics Informed Neural Networks is 
to learn both 
the network parameters $\vec{\theta}$ and 
the physics parameters $\vec{\vartheta}$
by simultaneously 
training a neural network (\ref{neuralnetwork}) and
solving an underlying physics equation (\ref{oscillator}).
We convert both into a single optimization problem 
that minimizes the loss function $L$ 
to learn the parameters 
$\vec{\Theta} = \{ \, \vec{\theta},\vec{\vartheta} \, \}$,
\beq
   L (\vec{\Theta}; t) 
= (\,1-\varepsilon\,) \,
   L_{\rm{data}} %(\vec{\theta}, \vec{\vartheta}; t) 
+      \varepsilon \, 
   L_{\rm{phys}} %(\vec{\theta}, \vec{\vartheta}; t)\,.
   \rightarrow \mbox{min} \,.
\label{loss_PINN}
\eeq
The first term, the data loss $L_{\rm{data}}$, penalizes the error between model $\mat{x}(\mat{t})$ and data $\hat{\mat{x}}$,
the second term, the physics loss $L_{\rm{phys}}$, penalizes the physics residual
$\mat{r}$.
Since both terms can be of a different order of magnitude, we weight them with the weighting coefficients
$(\,1-\varepsilon\,)$ and $\varepsilon$. 
For the first term, we use the mean square error, the $L_2$-norm of the difference between model and data, 
$||\, \mat{x}(\mat{t}) - \hat{\mat{x}}\,||^2$, 
divided by the number of training points $n_{\rm{trn}}$ similar to Section \ref{NN},
\beq
  L_{\rm{data}} (\vec{\Theta}; t)
= \frac{1}{n_{\rm{trn}}} \sum_{i=1}^{n_{\rm{trn}}}
|| \, x(t_i) - \hat{x}_i \, ||^2 \,.
\label{loss_data_PINN}
\eeq
For the second term, we use the $L_2$-norm of the residual,
$||\, \mat{r} \,||^2$,
divided by the total number of sampling points $n_{\rm{smp}}$,
\beq
  L_{\rm{phys}} (\vec{\Theta}; t)
= \frac{1}{n_{\rm{smp}}} \sum_{i=1}^{n_{\rm{smp}}}
|| \, r(t_i)  \, ||^2
= \frac{1}{n_{\rm{smp}}} \sum_{i=1}^{n_{\rm{smp}}}
|| \,        \ddot{x}(t_i) 
+ \frac{c}{m} \dot{x}(t_i) 
+ \frac{k}{m} \,  (x (t_i) - x_0 )\, ||^2 \,.
\label{loss_physics_PINN}
\eeq
A key step in calculating the physical loss function (\ref{loss_physics_PINN}) is to compute the time derivatives of the model solution $x$,
$\dot{x} = \sca{d}x/\sca{d}t$ and
$\ddot{x} = \sca{d}\dot{x}/\sca{d}t$,
which we address using automatic differentiation. Automatic differentiation is part of various deep learning frameworks which makes it convenient for Physics Informed Neural Networks to evaluate derivatives. 
We minimize the overall loss function (\ref{loss_PINN})
to optimize the network parameters $\vec{\theta}$, 
the network weights $\mat{W}_k$ and biases $\mat{b}_k$,
and the physics parameters $\vec{\vartheta}$,
the viscous damping $c$, stiffness $k$, and offset $x_0$,
\beq
\vec{\theta} = \{ \mat{W}_k, \mat{b}_k \}
\quad \mbox{and} \quad 
\vec{\vartheta} = \{ c, k, x_0 \}\,,
\eeq
again using the ADAM optimizer, 
an adaptive algorithm for gradient-based first-order optimization. 
%%%%%%%%%%%%%%%%%%%%%%%%%%%%%%%%%%%%%%%%%%%%%%%%%%%%%%%%%%%%%%%%%%%
\subsection{Self Adaptive Physics Informed Neural Networks}\label{SAPINN}
%%%%%%%%%%%%%%%%%%%%%%%%%%%%%%%%%%%%%%%%%%%%%%%%%%%%%%%%%%%%%%%%%%%
\noindent
The objective of Self Adaptive Physics Informed Neural Networks is to 
train a neural network (\ref{neuralnetwork}) and
solve an underlying physics equation (\ref{oscillator})
similar to Section \ref{PINN}, but now,
rather than prescribing the weighting coefficient $\varepsilon$ between data and physics,
we learn it as a function of time $t$, along with the other parameters,
$\vec{\Theta} = \{ \, \vec{\theta},\vec{\vartheta}, \epsilon \, \}$ \cite{mcclenny20}.
We adopt the same loss function $L$ as in Section \ref{PINN}, 
\beq
L (\vec{\Theta}; t) 
= [\,1-\varepsilon(t)\,] \,L_{\rm{data}} + \varepsilon(t) \, L_{\rm{phys}} \,,
\label{loss_sa}
\eeq
in terms of the data loss, 
the weighted error between model $\mat{x}$ and data $\mat{x}_{\rm{data}}$, 
\beq
  L_{\rm{data}} (\vec{\Theta}; t)
= \frac{1}{n_{\rm{trn}}} \sum_{i=1}^{n_{\rm{trn}}}
|| \, x(t_i) - \hat{x}_i \, ||^2 \,,
\label{loss1_sa}
\eeq
and the physics loss, 
the weighted norm of the physics residual $\mat{r}$,
\beq
  L_{\rm{phys}} (\vec{\Theta}; t)
= \frac{1}{n_{\rm{smp}}} \sum_{i=1}^{n_{\rm{smp}}}
|| \, r(t_i)  \, ||^2
= \frac{1}{n_{\rm{smp}}} \sum_{i=1}^{n_{\rm{smp}}}
|| \,        \ddot{x}(t_i) 
+ \frac{c}{m} \dot{x}(t_i) 
+ \frac{k}{m} \,  (x (t_i) - x_0 )\, ||^2 \,.
\label{loss2_sa}
\eeq
We minimize the overall loss function (\ref{loss_sa}) to optimize 
the network parameters, 
$\mat{W}_k$ and $\mat{b}_k$,
the physics parameters, $c$, $k$, and $x_0$, and
the weighting term $\varepsilon(t)$,
\beq
\vec{\theta} = \{ \mat{W}_k, \mat{b}_k \}
\quad \mbox{and} \quad 
\vec{\vartheta} = \{ m, c, k, x_0 \}
\quad \mbox{and} \quad 
\varepsilon(t) \,.
\eeq
The dynamics of the weighting term $\varepsilon(t)$ provide insight into the relative importance of data loss $L_{\rm{data}}$ and physics loss $L_{\rm{phys}}$ and into the change of both terms in time.
%%%%%%%%%%%%%%%%%%%%%%%%%%%%%%%%%%%%%%%%%%%%%%%%%%%%%%%%%%%%%%%%%%%
\subsection{Comparison of Neural Network models}\label{NNEX}
\begin{figure}[h]
\centering
\includegraphics[width=0.8\linewidth]{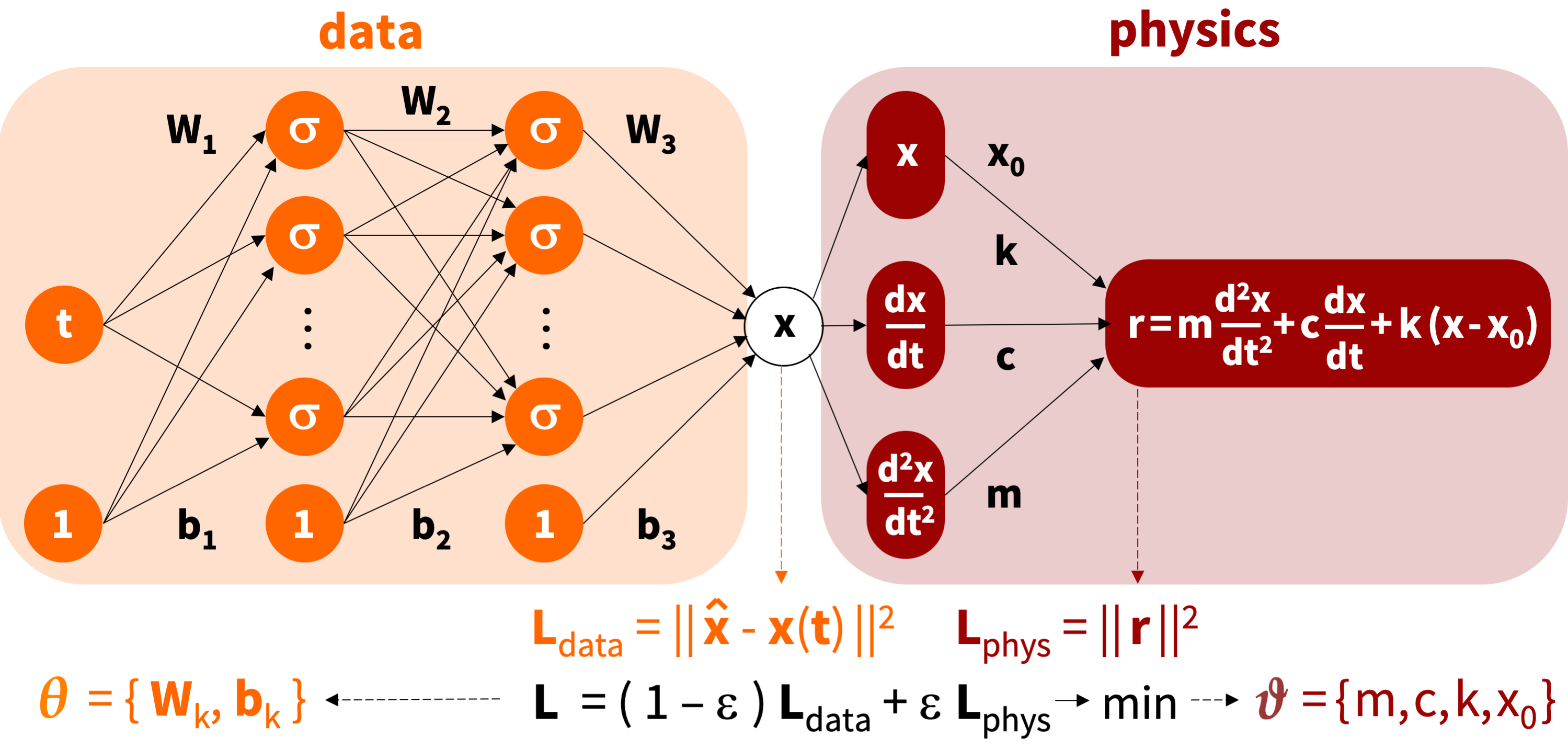}
\caption{{\bf{\sffamily{Neural Network modeling.}}} 
Neural Networks minimize a loss function $L\rightarrow\rm{min}$ 
that could consist of two terms, 
a data loss $L_{\rm{data}}$, 
the error between data and model
$|| \, \hat{\mat{x}} - \mat{x}(\mat{t}) \, ||$, and
a physics loss $L_{\rm{phys}}$,
the physics residual
$|| \, \mat{r} \, ||$,
to learn the network parameters 
$\vec{\theta} = \{ \mat{W}_k, \mat{b}_k \}$,
the physics parameters
$\vec{\vartheta} = \{ c, k, x_0 \}$,
and possibly even the dynamic weighting term 
$\varepsilon(t)$ between $L_{\rm{data}}$ and $L_{\rm{phys}}$.}
\label{figNN}
\end{figure}

%%%%%%%%%%%%%%%%%%%%%%%%%%%%%%%%%%%%%%%%%%%%%%%%%%%%%%%%%%%%%%%%%%%
\noindent
Figure \ref{figNN} illustrates the features the Neural Network models of this section. 
The classical Neural Networks from Section \ref{NN}
minimize a loss function, $L\rightarrow\rm{min}$,
that characterizes the data loss $L_{\rm{data}}$, 
the error between data and model
$|| \, \hat{\mat{x}} - \mat{x}(\mat{t}) \, ||$
shown on the left.
The Physics Informed Neural Networks from Section \ref{PINN}
build additional physics loss tersm $L_{\rm{phys}}$
into the loss function $L$, for example, 
the physics residual
$|| \, \mat{r} \, ||$
shown on the right. 
In a forward problem, Neural Networks learn the network parameters 
$\vec{\theta} = \{ \mat{W}_k, \mat{b}_k \}$.
In an inverse problem, Neural Networks learn the network parameters 
$\vec{\theta} = \{ \mat{W}_k, \mat{b}_k \}$.
the physics parameters
$\vec{\vartheta} = \{ c, k, x_0 \}$.
Self Adaptive Physics Informed Neural Networks from Section \ref{SAPINN}
also learn the dynamic weighting term 
$\varepsilon(t)$ between data and physics loss 
$L_{\rm{data}}$ and $L_{\rm{phys}}$.
%%%%%%%%%%%%%%%%%%%%%%%%%%%%%%%%%%%%%%%%%%%%%%%%%%%%%%%%%%%%%%%%%%%%%%%%
\\[4.pt]
%%%%%%%%%%%%%%%%%%%%%%%%%%%%%%%%%%%%%%%%%%%%%%%%%%%%%%%%%%%%%%%%%%%%%%%%
%%%%%%%%%%%%%%%%%%%%%%%%%%%%%%%%%%%%%%%%%%%%%%%%%%%%%%%%%%%%%%%%%%%
\begin{figure}[h]
\centering
\includegraphics[width=0.8\linewidth]{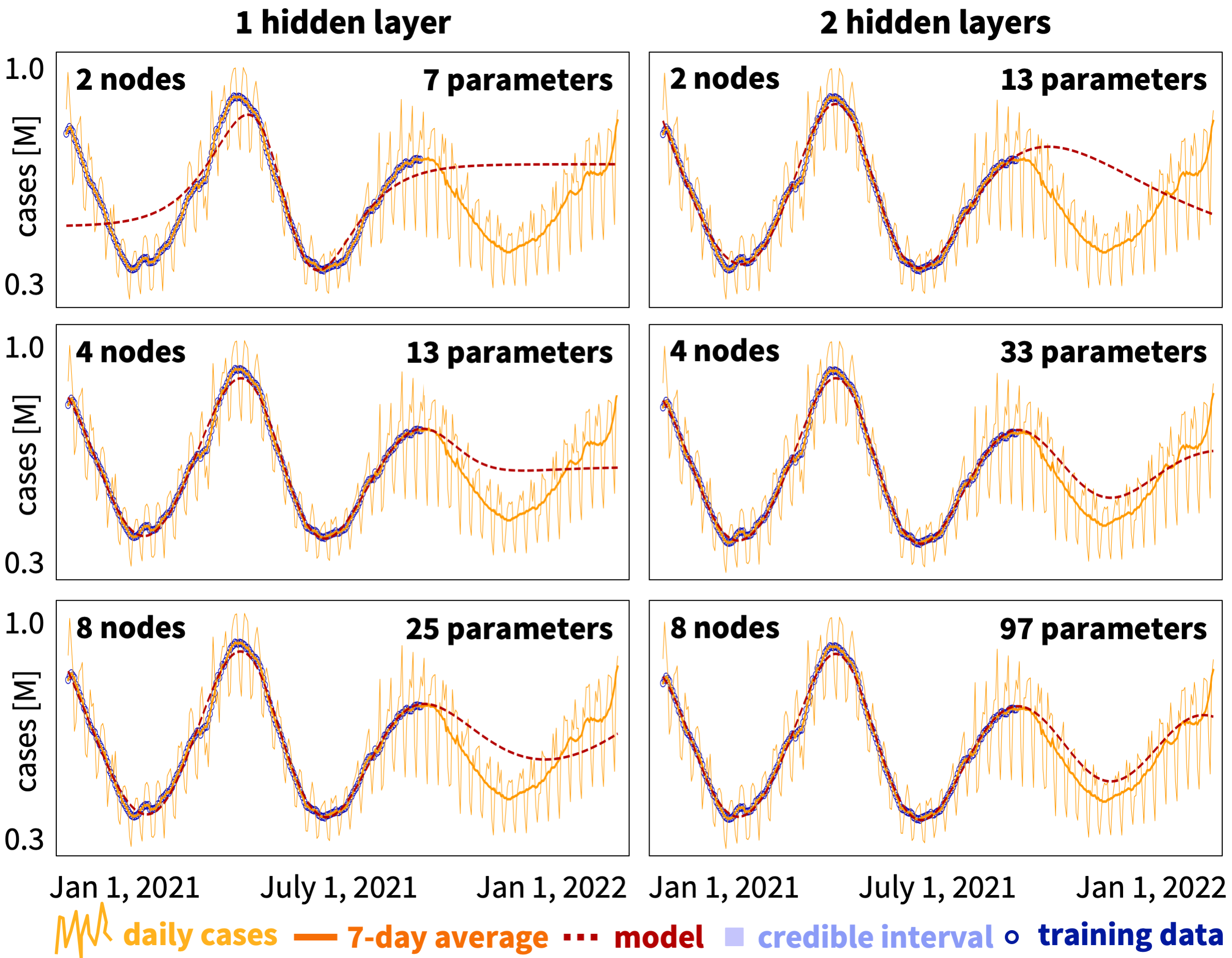}
\caption{{\bf{\sffamily{Neural Network modeling.
Grid search for Physics Informed Neural Network.}}}
The performance and number of parameters of the Physics Informed Neural Network
increases with increasing number of hidden layers (left to right) and increasing number of nodes per layer (top to bottom). 
For both one and two hidden layers, the performance does not increase markedly beyond eight nodes (bottom row).
Throughout this manuscript, we use a Neural Network with two hidden layers and 32 nodes. 
Thin yellow lines indicate daily new case data, thick orange lines are their seven-day average, dashed red lines are the physics-based model, light blue areas are its credible intervals, blue dots are the training data.}
\label{figgrid}
\end{figure}
%%%%%%%%%%%%%%%%%%%%%%%%%%%%%%%%%%%%%%%%%%%%%%%%%%%%%%%%%%%%%%%%%%%
Figure \ref{figgrid} illustrates the result of a grid search for the Physics Informed Neural Network. Here we show the simulation with one and two hidden layers, left and right columns, with two, four, and eight nodes each, from top to bottom. The number of network weights $\mat{W}_k$ and biases $\vec{b}_k$ increases with increasing number of layers and nodes. The one-hidden-layer models have 7, 13, and 25 parameters and the two-hidden-layer models have 13, 33, and 97 parameters. 
The six graphs confirm our intuition that the performance of the Physics Informed Neural Network increases with increasing number of hidden layers, from left to right, and increasing number of nodes per layer, from top to bottom. 
For both models, we also performed simulations with 16 and 32 nodes, but the performance did not increase markedly beyond the eight-node models.
Figure \ref{figgrid} suggests that the one-hidden-layer models perform poorly, independent of the number of nodes. 
For each additional hidden layer, the number of parameters increases with the number of nodes squared and the risk of overfitting increases.
The two-hidden-layer models seem to be a reasonable compromise between underfitting and overfitting: They perform well, even with only eight nodes and 97 parameters. Throughout this manuscript, we use Neural Networks with two hidden layers and 32 nodes. \\[4.pt]
%%%%%%%%%%%%%%%%%%%%%%%%%%%%%%%%%%%%%%%%%%%%%%%%%%%%%%%%%%%%%%%%%%%
\noindent
Figure \ref{fig04} compares the classical Neural Network with Physics Informed Neural Networks with varying weighting coefficient $\varepsilon$.
The thin yellow lines indicate daily new case data, the thick orange lines are their seven-day average, the blue dots are the training data, here for 225 out of 365 days,
the dashed red lines are the simulation with the physics-based model. 
In Physics Informed Neural Networks, the loss function,
$L = (\,1-\varepsilon\,) \,L_{\rm{data}} + \varepsilon \, L_{\rm{phys}}$,
is a weighted average of the data-based mean square error, $L_{\rm{data}}$, and the physics-based residual, $L_{\rm{phys}}$.
%%%%%%%%%%%%%%%%%%%%%%%%%%%%%%%%%%%%%%%%%%%%%%%%%%%%%%%%%%%%%%%%%%%
\begin{figure}[h]
\centering
\includegraphics[width=0.8\linewidth]{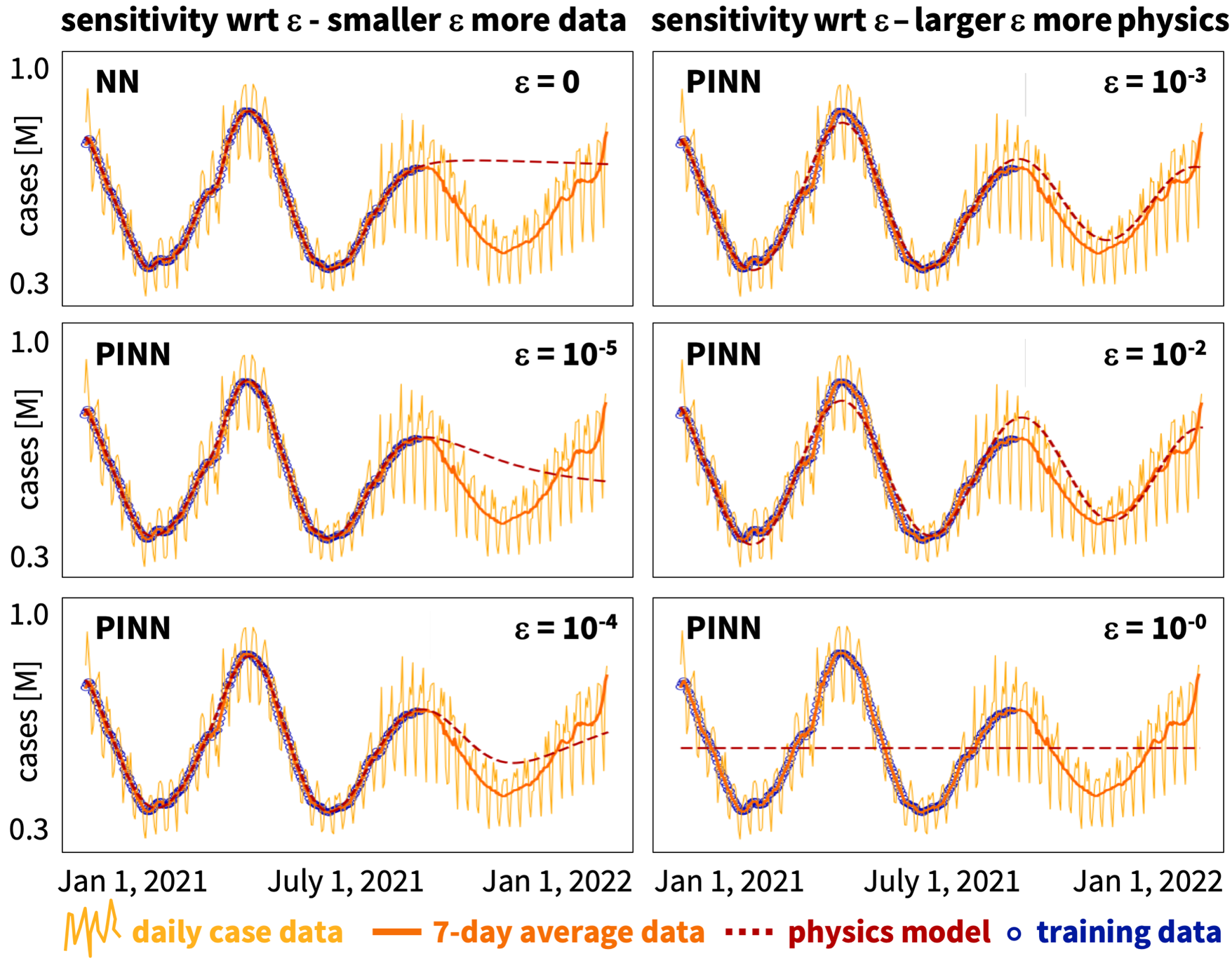}
\caption{{\bf{\sffamily{Neural Network modeling.
Classical Neural Network vs Physics Informed Neural Networks with 
varying weighting coefficients\,$\vec{\varepsilon}$}}.} 
In Physics Informed Neural Networks, the loss function,
$L = (\,1-\varepsilon\,) \,L_{\rm{data}} + \varepsilon \, L_{\rm{phys}}$,
is a weighted average of the data-based mean square error, $L_{\rm{data}}$, and the physics-based residual, $L_{\rm{phys}}$. 
The special case of $\varepsilon=0$ represents a classical Neural Network (top left).
With increasing weighting coefficients, 
$\varepsilon = [\, 10^{-\infty}, 10^{-5}, 10^{-4}, 10^{-3}, 10^{-2}, 10^{-0}] $,
the influence of the data fit decreases and 
the influence of the physics increases (top left to bottom right). 
Thin yellow lines indicate daily new case data, thick orange lines are their seven-day average, dashed red lines are the physics-based model, light blue areas are its credible intervals, blue dots are the training data, here for 225 out of 365 days.}
\label{fig04}
\end{figure}
%%%%%%%%%%%%%%%%%%%%%%%%%%%%%%%%%%%%%%%%%%%%%%%%%%%%%%%%%%%%%%%%%%%
For the special case of $\varepsilon=0$ in the top left, the loss function degenerates to the loss function of the classical Neural Network 
$L \rightarrow L_{\rm{data}}$ from Section \ref{NN}.
The classical Neural Network provides a good approximation of the training data during the first 225 days of the year, but fails to predict the behavior outside the training window. During the remaining 140 days, the classical Neural Network simply continues the learned trend, which, on day 225, suggests a plateau with constant case numbers for the remaining part of the year. 
%%%
For positive weighting coefficients $\varepsilon>0$, the loss function accounts for both data loss $L_{\rm{data}}$ and physics loss $L_{\rm{phys}}$.
With increasing weighting coefficients 
$\varepsilon = [\, 10^{-\infty}, 10^{-5}, 10^{-4}, 10^{-3}, 10^{-2}, 10^{-0}] $,
from top left to bottom right,
the influence of the data loss $L_{\rm{data}}$ decreases and 
the influence of the physics loss $L_{\rm{phys}}$ increases. 
In other words, 
the smaller the weighting coefficient $\varepsilon$, the better the fit of the data; 
the larger the weighting coefficient $\varepsilon$, the better the fit of the physics.
%%%
For the special case of $\varepsilon=1$ in the bottom right, the loss function degenerates to the loss function of the physical model
$L \rightarrow L_{\rm{phys}}$ from Section \ref{physics}. The network solves the physics of the damped harmonic oscillator without accounting for any data, and converges to the trivial solution with a zero amplitude. 
%%%%%%%%%%%%%%%%%%%%%%%%%%%%%%%%%%%%%%%%%%%%%%%%%%%%%%%%%%%%%%%%%%%
\begin{figure}[h]
\centering
\includegraphics[width=0.85\linewidth]{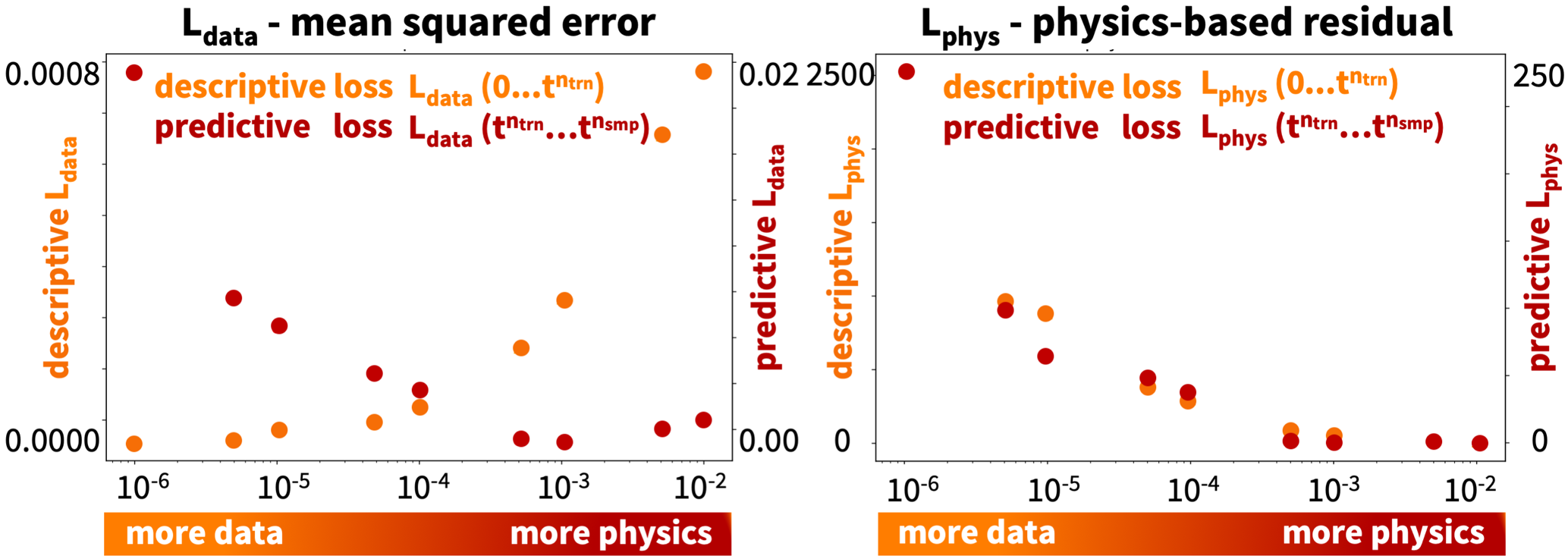}
\caption{{\bf{\sffamily{Neural Network modeling.
Sensitivity analysis for Physics Informed Neural Networks with 
varying weighting coefficients\,$\vec{\varepsilon}$}}.} 
In Physics Informed Neural Networks, the loss function,
$L = (\,1-\varepsilon\,) \,L_{\rm{data}} + \varepsilon \, L_{\rm{phys}}$,
is a weighted average of the data-based mean square error, $L_{\rm{data}}$, and the physics-based residual, $L_{\rm{phys}}$. 
With increasing weighting coefficients $\varepsilon$,
the model emphasizes less on the data fit and more on the physics fit;
the predictive data loss, descriptive physics loss, and predictive physics loss decrease, while the descriptive data loss increases.
Orange dots indicate the descriptive losses 
$L_{\rm{data}}$ and $L_{\rm{phys}}$ 
for the first 255 days; 
red dots indicate the predictive losses 
$L_{\rm{data}}$ and $L_{\rm{phys}}$ 
for the following 140 days.}
%where $\rm{n}_{\rm{trn}}$ = 225 days and $\rm{n}_{\rm{smp}}$ = 365 days.}
\label{fig05}
\end{figure}\\[6.pt]
%%%%%%%%%%%%%%%%%%%%%%%%%%%%%%%%%%%%%%%%%%%%%%%%%%%%%%%%%%%%%%%%%%%
\noindent
Figure \ref{fig05} quantifies the contributions to the loss function for 
Physics Informed Neural Networks with weighting coefficients $\varepsilon$ varying from 10$^{-6}$ to 10$^{-2}$.
In Physics Informed Neural Networks, the loss function,
$L = (\,1-\varepsilon\,) \,L_{\rm{data}} + \varepsilon \, L_{\rm{phys}}$,
is a weighted average of 
the data-based mean square error, $L_{\rm{data}}$, summarized in the left panel, and 
the physics-based residual, $L_{\rm{phys}}$, summarized in the right panel.
Each panel separately reports the descriptive loss, the sum of the daily errors in the training window as orange dots,
\[
  L_{\rm{data}} (\vec{\Theta}; t)
= \frac{1}{255} \sum_{i=1}^{255}
|| \, x(t_i) - \hat{x}_i \, ||^2 
 \qquad \mbox{and} \qquad
   L_{\rm{phys}} (\vec{\Theta}; t)
= \frac{1}{255} \sum_{i=1}^{255}
|| \, r(t_i)  \, ||^2 \,,
\]
and the predictive loss, the sum of the daily errors in the prediction window as red dots, 
\[
  L_{\rm{data}} (\vec{\Theta}; t)
= \frac{1}{140} \sum_{i=256}^{365}
|| \, x(t_i) - \hat{x}_i \, ||^2 
 \qquad \mbox{and} \qquad
  L_{\rm{phys}} (\vec{\Theta}; t)
= \frac{1}{140} \sum_{i=256}^{365}
|| \, r(t_i)  \, ||^2 \,.
\]
With increasing weighting coefficients $\varepsilon$,
the model emphasizes less on the data fit and more on the physics fit.
The predictive data loss, descriptive physics loss, and predictive physics loss decrease, while the descriptive data loss increases.
The quantitative comparison of the four losses in Figure \ref{fig05} agrees with the qualitative observation in Figure~\ref{fig04}:
For smaller weighting coefficients $\varepsilon$, the red dashed lines of the model and orange lines of the seven-day case data average are in close proximity during the first 225 days of the descriptive regime, but deviate notably during the remaining 140 days of the predictive regime. 
For larger weighting coefficients $\varepsilon$, the red and orange lines deviate more during the first 225 days of the descriptive regime, but are in close proximity during the remaining 140 days of the predictive regime.
This example highlights the importance of selecting an appropriate weighting coefficient $\varepsilon$, which can bias the result of a Physics Informed Neural Network either towards the data or towards the physics. 
An ideal weighting coefficient minimizes the sum of all four loss terms. From Figure \ref{fig05}, we can estimate a reasonable weighting coefficient on the order of $\varepsilon\approx 10^{-3}$. This value is of the same order of magnitude as the learned weighting term $\varepsilon(t)$ with mean and standard deviation 
of $\varepsilon = 0.007 \pm 0.003$ %[Mean $\pm$ SD]
for the self-adpative Physics Informed Neural Network from Section \ref{SAPINN}.
%%%%%%%%%%%%%%%%%%%%%%%%%%%%%%%%%%%%%%%%%%%%%%%%%%%%%%%%%%%%%%%%%%%
\begin{figure}
\centering
\includegraphics[width=0.8\linewidth]{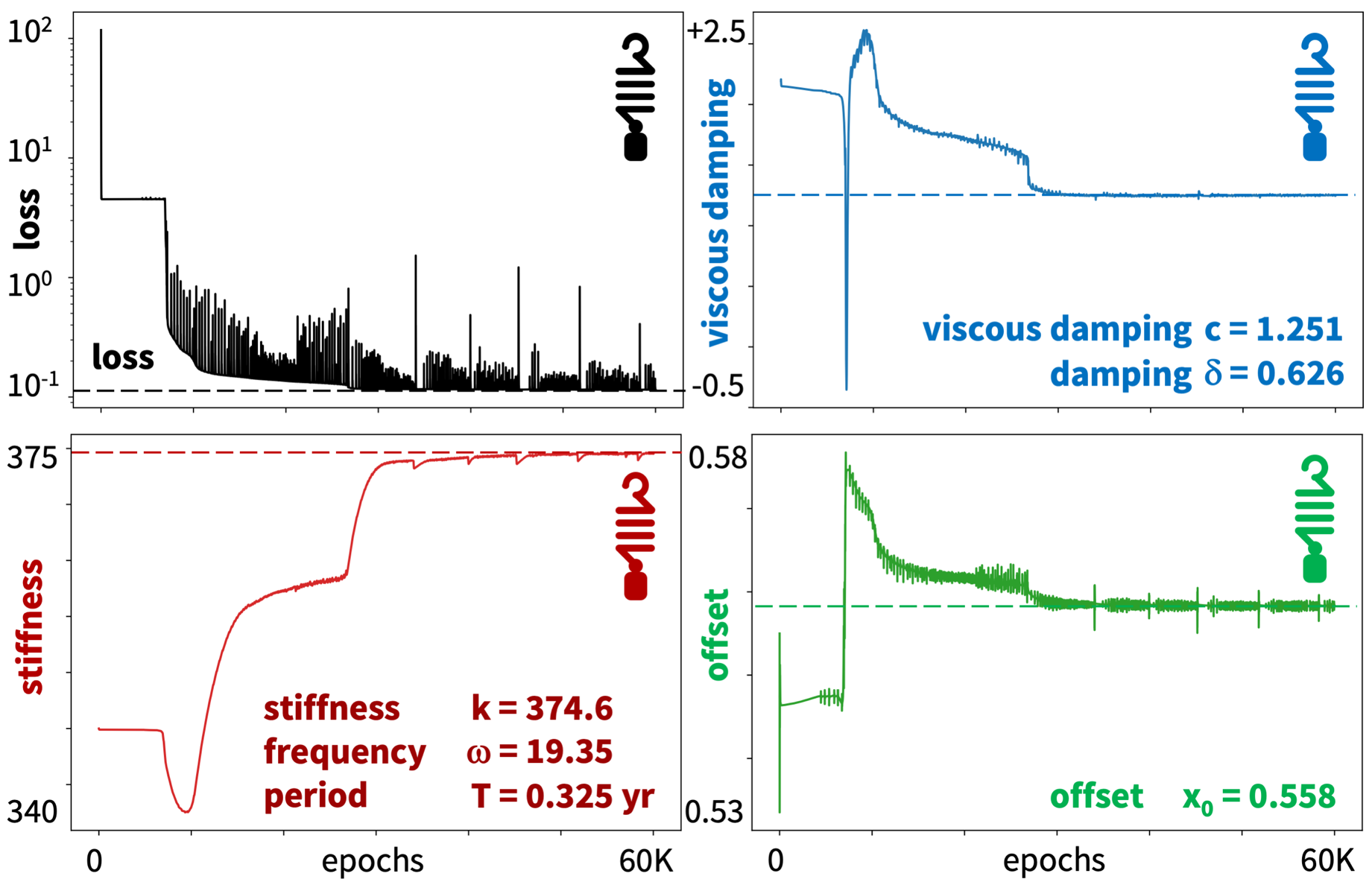}
\caption{{\bf{\sffamily{Physics Informed Neural Network. History of loss function $\vec{L}$ and physics parameters damping $\vec{c}$, stiffness $\vec{k}$,\,and\,offset\,$\vec{x_0}$.}}} Within 60K epochs, 
the physics parameters converge towards 
a viscous damping coefficient of $c=1.251$, 
a stiffness of $k=374.6$, and 
an offset of $x_0=0.558$.
This corresponds to 
a damping of $\delta=0.626$,
a frequency of $\omega = 19.35$,
and a period of $T=0.325$ years. 
Simulation with weighting coefficient of $\varepsilon=10^{-3}$ and training window of $n_{\rm{trn}}$ = 225 days.}
\label{fig06}
\end{figure}\\[6.pt]
%%%%%%%%%%%%%%%%%%%%%%%%%%%%%%%%%%%%%%%%%%%%%%%%%%%%%%%%%%%%%%%%%%%
Figure \ref{fig06} summarizes the history of the loss function $\vec{L}$ and of the physics parameters $\vec{\vartheta}$, the damping $c$, stiffness $k$, and offset $x_0$, for a Physics Informed Neural Network with a weighting coefficient of $\varepsilon = 10^{-3}$.
For the first 30K epochs, the loss function decreases while the physics parameters are still oscillating without clear trends. 
After 60K epochs, 
the physics parameters have converged towards 
a viscous damping coefficient of $c=1.251$, 
a stiffness of $k=374.6$, and 
an offset of $x_0=0.558$.
This corresponds to 
a damping of $\delta=0.626$,
a frequency of $\omega = 19.35$,
and a period of $T=0.325$ years.
This period of approximately four months agrees well with the reported daily new cases in Figure \ref{fig01} that display three pronounced waves throughout the year of 2021. 
%%%%%%%%%%%%%%%%%%%%%%%%%%%%%%%%%%%%%%%%%%%%%%%%%%%%%%%%%%%%%%%%%%%
\section{Bayesian Inference modeling}\label{BImodeling}
%%%%%%%%%%%%%%%%%%%%%%%%%%%%%%%%%%%%%%%%%%%%%%%%%%%%%%%%%%%%%%%%%%%
\noindent
The objective of Bayesian Inference is to estimate the posterior probability distribution of a set of the parameters
$\vec{\Theta}$, 
such that 
the statistics of the model $\mat{x}(\mat{t})$ agree with data ${\hat{\mat{x}}}$ \cite{bayes63} 
and possibly also satisfy the physics $\mat{r}$,
by maximizing the prior-weighted likelihood 
$P({\hat{\mat{x}}},\mat{r}|\vec{\Theta})$,
\beq
   P(\vec{\Theta}|{\hat{\mat{x}}},\mat{r}) 
= \frac
  {P({\hat{\mat{x}}},\mat{r}|\vec{\Theta}) \, P(\vec{\Theta})}
  {P({\hat{\mat{x}}})P(\mat{r})} 
= \frac
  {P({\hat{\mat{x}}}|\vec{\Theta}) \,P(\mat{r}|\vec{\Theta}) \, P(\vec{\Theta})}
  {P({\hat{\mat{x}}})P(\mat{r})} \rightarrow \rm{max}\,.
\label{BI_general}  
\eeq
Here, 
$P({\hat{\mat{x}}},\mat{r} | \vec{\Theta})$ 
is the likelihood of the data ${\hat{\mat{x}}}$  and
$P(\mat{r}|\vec{\Theta})$ 
the likelihood of the physics $\mat{r}$
for given fixed parameters $\vec{\Theta}$; 
$P(\vec{\Theta})$ is the prior, the probability distribution of the model parameters $\vec{\Theta}$; 
$P({\hat{\mat{x}}})$ and $P(\mat{r})$ are the marginal likelihood or evidence;
and 
$P(\vec{\Theta}|{\hat{\mat{x}}},\mat{r})$ is the posterior, the conditional probability of the parameters $\vec{\Theta}$ for the given data ${\hat{\mat{x}}}$ and physics $\mat{r}$.
%\\[4.pt]
%%% likelihood
The first likelihood function $P({\hat{\mat{x}}}|\vec{\Theta})$ evaluates the quality of fit between the model output $\mat{x}(\mat{t})$, here the simulated number of new cases for given parameters $\vec{\Theta}$, and the observed data $\hat{\mat{x}}$, here the reported number of new cases at every day $i$.
For the daily likelihood function $p_i(\hat{x}|\vec{\Theta})$, 
at each time point $t_i$, %$P(\hat{I}(t)|\vec{\vartheta})$, 
we select a normal distribution,
$\mathcal{N}(\mu,\sigma)$, where the mean $\mu$ is the model result $x(t_i)$ for the given parameter set $\vec{\Theta}$ and the likelihood width, the standard deviation $\sigma$, accounts for the observation error in the data $\hat{x}$.
The product $\prod_{i=0}^{\rm{n}}$
of all daily likelihood functions
$p_i(\hat{x}|\vec{\Theta})$, 
in our case throughout the ${\rm{n}}=365$ days of the year 2021,
defines the overall likelihood $P(\hat{\mat{x}}|\vec{\Theta})$,
\beq
  P(\hat{\mat{x}}|\vec{\Theta}) 
= \prod_{i=0}^{\rm{n}}
  p_i(\hat{x}|\vec{\Theta})
  \qquad \mbox{with} \qquad
  p_i(\hat{x}|\vec{\Theta})
%\sim \mathcal{N}(\mu = x(\vec{\Theta};t_i), \sigma )\,.
%  \quad \mbox{with} \quad
= \frac{1}{\sqrt{2\pi} \sigma} 
  {\rm{exp}} 
  \left(-\frac{||\hat{x}-x(t_i)||^2}{2 \sigma^2} \right).
  \label{BIlikelihood}
\eeq
The second likelihood function $P(\mat{r}|\vec{\Theta})$ evaluates how accurately the model output $\mat{x}(\mat{t})$, here the simulated number of new cases for given parameters $\vec{\Theta}$, satisfies the physics equation $\mat{r}(\mat{t}) = \mat{0}$ at every day $i$.
Again, we define the 
daily likelihood function $p_i(r|\vec{\Theta})$, 
at each time point $t_i$, %$P(\hat{I}(t)|\vec{\vartheta})$, 
through a normal distribution,
$\mathcal{N}(\mu,\sigma)$, where the mean $\mu$ is the model result of the physics equation $r(x(t_i))$ for the given parameter set $\vec{\Theta}$ and the likelihood width is the standard deviation $\sigma$.
The product $\prod_{i=0}^{\rm{n}}$
of all daily likelihood functions
$p_i(r|\vec{\Theta})$, 
defines the overall likelihood $P(\mat{r}|\vec{\Theta})$,
\beq
  P(\mat{r}|\vec{\Theta}) 
= \prod_{i=0}^{\rm{n}}
  p_i(r|\vec{\Theta})
  \qquad \mbox{with} \qquad
  p_i(r|\vec{\Theta})
%\sim \mathcal{N}(\mu = x(\vec{\Theta};t_i), \sigma )\,.
%  \quad \mbox{with} \quad
= \frac{1}{\sqrt{2\pi} \sigma} 
  {\rm{exp}} 
  \left(-\frac{||r(t_i)||^2}{2 \sigma^2} \right).
  \label{BIlikelihood_r}
\eeq
Similar to the previous section on Neural Network modeling, 
the set of model parameters
could consist of 
network parameters $\vec{\theta}$ and
physics parameters $\vec{\vartheta}$,
\beq
\vec{\Theta} = \{ \vec{\theta}, \vec{\vartheta} \}
\qquad \mbox{with} \qquad
\vec{\theta}= \{ \mat{W}_k, \mat{b}_k \} 
\qquad \mbox{and} \qquad
\vec{\vartheta}= \{ c, k, x_0 \},
\label{parametersBI_general}
\eeq
where
$\mat{W}_k$ and $\mat{b}_k$ are the network weights and biases 
from equation (\ref{neuralnetwork})
and
$c$, $k$, and $x_0$ are the physical damping, stiffness, and offset
from equation (\ref{oscillator}). 
We specify the Bayesian Inference (\ref{BI_general}) and the prior distributions of its parameters (\ref{parametersBI_general})
for classical Bayesian Inference in Section \ref{BI},
for Bayesian Neural Networks in Section \ref{BNN}, and
for Bayesian Physics Informed Neural Networks in Section \ref{BPINN}.
%%%%%%%%%%%%%%%%%%%%%%%%%%%%%%%%%%%%%%%%%%%%%%%%%%%%%%%%%%%%%%%%%%%
\subsection{Bayesian Inference}\label{BI}
%%%%%%%%%%%%%%%%%%%%%%%%%%%%%%%%%%%%%%%%%%%%%%%%%%%%%%%%%%%%%%%%%%%
\noindent
The objective of a classical Bayesian Inference is to estimate the posterior parameter distributions $P(\vec{\vartheta}|{\hat{\mat{x}}})$ of a set of physics parameters $\vec{\vartheta}$
such that the statistics of a physical model $\mat{x}(\mat{t})$ agree with data ${\hat{\mat{x}}}$.
For the physical model, we assume that the number of daily new cases $x(t)$ follows the physics of a damped harmonic oscillator from Section \ref{physics}, 
%and solve the the second order differential equation,
%\beq
$m \, \ddot{x} + c\, \dot{x} + k\, x =0$,
%\eeq
where 
$m$ is the mass,
$c$ is the viscous damping coefficient, and
$k$ is the stiffness. 
Without loss of generality, we set $m \equiv 1$.
We consider the underdamped case,
for which the damping ratio is smaller than one,
$c^2 < 4 \,k$.
For this case, following equation (\ref{analytical}), the analytical solution for the number of daily new cases is
\beq
  x(\vec{\vartheta};t) 
= x_0 
+ A_0 \cos \,
   ((\, k  - \mbox{$\frac{1}{2}$} c^2 )  \, t \, ) 
\, \exp (- \mbox{$\frac{1}{2}$} c \, t)  \,.
\label{BIxoft}
\eeq
The physics of the number of new cases are uniquely determined by three parameters,
$\vec{\vartheta} = \{ c, k, x_0 \}$,
the viscous damping $c$, the stiffness $k$, and the offset $x_0$.%\\[6.pt]
%%%%%%%%%%%%%%%%%%%%%%%%%%%%%%%%%%%%%%%%%%%%%%%%%%%%%%%%%%%%%%%%%%%%
%KEVIN: We estimate the model parameters by applying Bayesian inference with Markov Chain Monte Carlo sampling. This allows us to additionally learn the uncertainty of the constitutive parameters $\vec{\vartheta}$ on the basis of the reported case data $\hat{x}$. 
%%%%%%%%%%%%%%%%%%%%%%%%%%%%%%%%%%%%%%%%%%%%%%%%%%%%%%%%%%%%%%%%%%%
We now use Bayes' theorem, to estimate the posterior probability distribution of the physics parameters $\vec{\vartheta}$
such that the statistics of the model $\mat{x}(t)$ from equation (\ref{BIxoft}) agree with the reported case data ${\hat{\mat{x}}}$ in Figure \ref{fig01},
\beq
  P(\vec{\vartheta}|{\hat{\mat{x}}}) 
= \frac
  {P({\hat{\mat{x}}}|\vec{\vartheta}) \, P(\vec{\vartheta})}
  {P({\hat{\mat{x}}})} \,.
\label{BIposterior}  
\eeq
Here
$P({\hat{\mat{x}}}|\vec{\vartheta})$ is the likelihood, the conditional probability of the data ${\hat{\mat{x}}}$ for given fixed physics parameters $\vec{\vartheta}$; 
$P(\vec{\vartheta})$ is the prior, the probability distribution of the physics parameters $\vec{\vartheta}$; 
$P({\hat{\mat{x}}})$ is the marginal likelihood or evidence;
and 
$P(\vec{\vartheta}|{\hat{\mat{x}}})$ is the posterior, the conditional probability of the physics parameters $\vec{\vartheta}$ for the given data ${\hat{\mat{x}}}$.
%%%
For the likelihood $P({\hat{\mat{x}}}|\vec{\vartheta})$, the product of the individual point-wise likelihoods $p_i({\hat{x}}|\vec{\vartheta})$ according to equation (\ref{BIlikelihood}), we select a normal distribution ${\mathcal{N}}(\mu,\sigma)$, where the mean $\mu$ is the model result $x(t_i)$ for the given physics parameters $\vec{\vartheta}$ and the likelihood width $\sigma$ takes a half Cauchy distribution, $\sigma = \rm{halfCauchy}(\beta=1.0)$,
\beq
  P(\hat{\mat{x}}|\vec{\vartheta}) 
= \prod_{i=0}^{\rm{n}}
  p_i(\hat{x}|\vec{\vartheta})
  \qquad \mbox{with} \qquad
  p_i(\hat{x}|\vec{\vartheta})
%\sim \mathcal{N}(\mu = x(\vec{\Theta};t_i), \sigma )\,.
%  \quad \mbox{with} \quad
= \frac{1}{\sqrt{2\pi} \sigma} 
  {\rm{exp}} 
  \left(-\frac{||\hat{x}-x(t_i)||^2}{2 \sigma^2} \right).
  \label{BIlikelihood_BI}
\eeq
%%%%%%%%%%%%%%%%%%%%%%%%%%%%%%%%%%%%%%%%%%%%%%%%%%%%%%%%%%%%%%%%%%%
% prior probability distributions
%%%%%%%%%%%%%%%%%%%%%%%%%%%%%%%%%%%%%%%%%%%%%%%%%%%%%%%%%%%%%%%%%%%
For the prior probability distributions
$P ( {\vec{\vartheta}})$, 
we select independent weakly informed priors with
log-normal distributions for the three physics parameters, 
$\vec{\vartheta}=\{ c, k, x_0 \}$,
the viscous damping $c$, 
the stiffness $k$, and
the offset $x_0$,
\beq
\begin{array}{@{\hspace*{0.0cm}}l@{\hspace*{0.0cm}}l}
  P ( {\vec{\vartheta}})
= P ( c, k, x_0 )
  \qquad \mbox{with} \qquad   
\begin{array}{ @{\hspace*{0.0cm}}l@{\hspace*{0.1cm}}
              c@{\hspace*{0.1cm}}
              l@{\hspace*{0.1cm}}l}                          
%\log (A_0) &\sim& {\mathcal{N}}(\mu=\log(0.5), & \sigma=0.5)    \\
\log (c)   &\sim& {\mathcal{N}}(\mu=\log(2.2), & \sigma=0.5)    \\
\log (k)   &\sim& {\mathcal{N}}(\mu=\log(350), & \sigma=0.5)    \\
\log (x_0) &\sim& {\mathcal{N}}(\mu=\log(0.56),& \sigma=0.5).
\end{array}
\end{array}
\label{BIpriors}
\eeq
Using the case data from Figure \ref{fig01}, we approximately infer the posterior distributions from Bayes theorem (\ref{BIposterior}) by employing a NO-U-Turn sampler \cite{hoffman14} implementation of the Python package PyMC3 \cite{salvatier16}. We use two chains with the first 2000 samples to tune the sampler and the subsequent 4000 samples to estimate the set of parameters ${\vec{\vartheta}}$, and employ a target acceptance rate of 0.85.
%%%%%%%%%%%%%%%%%%%%%%%%%%%%%%%%%%%%%%%%%%%%%%%%%%%%%%%%%%%%%%%%%%%
\subsection{Bayesian Neural Networks}\label{BNN}
%%%%%%%%%%%%%%%%%%%%%%%%%%%%%%%%%%%%%%%%%%%%%%%%%%%%%%%%%%%%%%%%%%%
\noindent
The objective of Bayesian Neural Networks is to estimate the posterior parameter distributions $P(\vec{\theta}|{\hat{\mat{x}}})$ of a set of network parameters $\vec{\theta}$, without any prior information about the underlying physics,
such that the statistics of the output of the neural network $\mat{x}(\mat{t})$ agree with data ${\hat{\mat{x}}}$.
For the neural network, we use the fully-connected feed-forward neural network composed of multiple hidden layers of Section \ref{neural_networks}
to approximate the output, the number of daily new cases $\mat{x}$,
from the input, the time coordinates $\mat{t}$.
We denote the hidden variables of the $k_{\rm{th}}$ hidden layer as $\mat{z}^k$
and summarize the neural network, here for two hidden layers, as 
\beq
\begin{array}{l@{\hspace*{0.2cm}}c@{\hspace*{0.2cm}}
              l@{\hspace*{0.2cm}}l@{\hspace*{0.2cm}}
              l@{\hspace*{0.2cm}}l}
\mat{z}_0 &=&            &                       &  &\mat{t} \\
\mat{z}_1 &=& \sigma \,( &\mat{W}_1 \, \mat{z}_0 &+ &\mat{b}_1\,)  \\
\mat{z}_2 &=& \sigma \,( &\mat{W}_2 \, \mat{z}_1 &+ &\mat{b}_2\,)  \\
\mat{z}_3 &=&            &\mat{W}_3 \, \mat{z}_2 &+ &\mat{b}_3\,, 
\end{array}
\label{BIneuralnetwork}
\eeq
where the output of the last layer approximates the true solution,
$\mat{x} \approx \mat{z}_3$.
Similar to the previous sections,
$\sigma \,(\circ)$ denotes the non-linear activation function for which we select a hyperbolic tangent function, $\sigma \,(\circ)= \tanh\,(\circ)$,
and 
$\mat{W}_k$ and $\mat{b}_k$ are the network weights and biases of the $k^{th}$ layer.
We use Bayes' theorem, to estimate the posterior probability distribution of the network parameters $\vec{\theta}$
such that the statistics of the neural network $\mat{x}\approx \mat{z}_3$ from equation (\ref{BIneuralnetwork}) agree with the reported case data ${\hat{\mat{x}}}$ in Figure \ref{fig01},
\beq
   P(\vec{\theta}|{\hat{\mat{x}}}) 
= \frac
  {P({\hat{\mat{x}}}|\vec{\theta}) \, P(\vec{\theta})}
  {P({\hat{\mat{x}}})} \,.
\label{BNNposterior}  
\eeq
Here
$P({\hat{\mat{x}}}|\vec{\theta})$ is the likelihood, the conditional probability of the data $\hat{\mat{x}}$ for given fixed parameters $\vec{\theta}$; 
$P(\vec{\theta})$ is the prior, the probability distribution of the network parameters $\vec{\theta}$; 
$P(\hat{\mat{x}})$ is the marginal likelihood or evidence;
and 
$P(\vec{\theta}|\hat{\mat{x}})$ is the posterior, the conditional probability of the parameters $\vec{\theta}$ for the given data $\hat{\mat{x}}$. %\\[6.pt]
For the likelihood $P({\hat{\mat{x}}}|\vec{\theta})$, the product of the individual daily point-wise likelihoods $p_i({\hat{x}}|\vec{\theta})$ according to equation (\ref{BIlikelihood}), we select a normal distribution ${\mathcal{N}}(\mu,\sigma)$, where the mean $\mu$ is the model result $x(t_i)$ for the given network parameters $\vec{\theta}$ and the likelihood width $\sigma$
is the standard deviation with $\sigma = 0.05$,
\beq
  P(\hat{\mat{x}}|\vec{\theta}) 
= \prod_{i=0}^{\rm{n}}
  p_i(\hat{x}|\vec{\theta})
  \qquad \mbox{with} \qquad
  p_i(\hat{x}|\vec{\theta})
%\sim \mathcal{N}(\mu = x(\vec{\Theta};t_i), \sigma )\,.
%  \quad \mbox{with} \quad
= \frac{1}{\sqrt{2\pi} \sigma} 
  {\rm{exp}} 
  \left(-\frac{||\hat{x}-x(t_i)||^2}{2 \sigma^2} \right).
  \label{BIlikelihood_BNN}
\eeq
For the prior probability distributions
$P ( {\vec{\theta}})$, 
we select independent weakly informed priors with
normal distributions with a zero-mean for the two sets of network parameters, 
$\vec{\theta}=\{ \mat{W}_k, \mat{b}_k \}$,
the network weights $\mat{W}_k$ and biases  $\mat{b}_k$ \cite{yang21}, 
\beq
\begin{array}{@{\hspace*{0.0cm}}l@{\hspace*{0.0cm}}l}
  P ( {\vec{\theta}})
= P (  \mat{W}_k, \mat{b}_k )
  \qquad \mbox{with} \qquad   
\begin{array}{ @{\hspace*{0.0cm}}r@{\hspace*{0.1cm}}
              c@{\hspace*{0.1cm}}
              l@{\hspace*{0.1cm}}l}                          
\mat{W}_k &\sim& {\mathcal{N}}(\mu=0.0, & \sigma=2.0)    \\
\mat{b}_k &\sim& {\mathcal{N}}(\mu=0.0, & \sigma=2.0) \,. 
\end{array}
\end{array}
\label{BIpriors}
\eeq
Using the case data from Figure \ref{fig01}, we approximately infer the posterior distributions from Bayes theorem (\ref{BNNposterior}) by employing a Hamiltonian Monte Carlo sampling \cite{yang21} using Tensorflow-Probability \cite{tensorflow20}.
We use the first 3000 samples to tune the sampler and the subsequent 3000 samples to estimate the set of parameters ${\vec{\theta}}$.
%%%%%%%%%%%%%%%%%%%%%%%%%%%%%%%%%%%%%%%%%%%%%%%%%%%%%%%%%%%%%%%%%%%
\subsection{Bayesian Physics Informed Neural Networks}\label{BPINN}
%%%%%%%%%%%%%%%%%%%%%%%%%%%%%%%%%%%%%%%%%%%%%%%%%%%%%%%%%%%%%%%%%%%
\noindent
The objective of Bayesian Physics Informed Neural Networks is to estimate the posterior parameter distributions $P(\vec{\Theta}|{\hat{\mat{x}}},\mat{r})$ of both network and physics parameters, $\vec{\Theta}=\{\,\vec{\theta},\vec{\vartheta}\,\}$,
such that the statistics of Physics Informed Neural Network $\mat{x}(\mat{t})$ agree with data ${\hat{\mat{x}}}$ and satisfy the physics equation $\mat{r}$.
As such, the Bayesian Physics Informed Neural Network integrates the Physics Informed Neural Network from Section \ref{PINN} into a Bayesian Inference.
We use Bayes' theorem, to estimate the posterior probability distribution of the  parameters 
such that the statistics of the neural network $\mat{x}\approx \mat{z}_3$ from equation (\ref{neuralnetwork}) agree with the reported case data ${\hat{\mat{x}}}$ in Figure \ref{fig01},
\beq
   P(\vec{\Theta}|{\hat{\mat{x}}},\mat{r}) 
= \frac
  {P({\hat{\mat{x}}},\mat{r}|\vec{\Theta}) \, P(\vec{\Theta})}
  {P({\hat{\mat{x}}}) \, P({\mat{r}})}
= \frac
  {P({\hat{\mat{x}}}|\vec{\Theta}) \,P({\mat{r}}|\vec{\Theta}) 
\, P(\vec{\theta}) \, P(\vec{\vartheta})}
  {P({\hat{\mat{x}}}) \, P({\mat{r}})} \,.
\label{BPINNposterior}  
\eeq
Here
 $P({\hat{\mat{x}}},\mat{r}|\vec{\Theta})
= P({\hat{\mat{x}}}|\vec{\Theta}) 
\,P({\mat{r}}|\vec{\Theta})$ 
are the two likelihood functions, 
the conditional probabilities of data $\hat{\mat{x}}$ and physics $\mat{r}$ for given parameters $\vec{\Theta}$; 
$P(\vec{\Theta})=P(\vec{\theta}) \, P(\vec{\vartheta})$ are the priors, the probability distribution of the network and model parameters 
$\vec{\theta}$ and $\vec{\vartheta}$; 
$P(\hat{\mat{x}})\,P(\mat{r})$ is the marginal likelihood or evidence;
and 
$P(\vec{\Theta}|\hat{\mat{x}},\mat{r})$ is the posterior, the conditional probability of the parameters $\vec{\Theta}$ for given data $\hat{\mat{x}}$ and physics $\mat{r}$. 
%%%
The first likelihood function $P({\hat{\mat{x}}}|\vec{\Theta})$ evaluates the quality of fit between the model output $\mat{x}(\mat{t})$ and the observed data $\hat{\mat{x}}$ at every day $i$
as a product of the daily likelihood functions $p_i(\hat{x}|\vec{\Theta})$,
for which we select a normal distribution,
\beq
  P(\hat{\mat{x}}|\vec{\Theta}) 
= \prod_{i=0}^{\rm{n}}
  p_i(\hat{x}|\vec{\Theta})
  \qquad \mbox{with} \qquad
  p_i(\hat{x}|\vec{\Theta})
%\sim \mathcal{N}(\mu = x(\vec{\Theta};t_i), \sigma )\,.
%  \quad \mbox{with} \quad
= \frac{1}{\sqrt{2\pi} \sigma} 
  {\rm{exp}} 
  \left(-\frac{||\hat{x}-x(t_i)||^2}{2 \sigma^2} \right).
  \label{BIlikelihood}
\eeq
The second likelihood function $P(\mat{r}|\vec{\Theta})$ evaluates how accurately the model output $\mat{x}(\mat{t})$ satisfies the physics equation $\mat{r}= \mat{0}$ at every day $i$
as a product of the daily likelihood functions $p_i(r|\vec{\Theta})$,
for which we also assume a normal distribution,
\beq
  P(\mat{r}|\vec{\Theta}) 
= \prod_{i=0}^{\rm{n}}
  p_i(r|\vec{\Theta})
  \qquad \mbox{with} \qquad
  p_i(r|\vec{\Theta})
%\sim \mathcal{N}(\mu = x(\vec{\Theta};t_i), \sigma )\,.
%  \quad \mbox{with} \quad
= \frac{1}{\sqrt{2\pi} \sigma} 
  {\rm{exp}} 
  \left(-\frac{||r(t_i)||^2}{2 \sigma^2} \right).
  \label{BIlikelihood_r}
\eeq
For the prior probability distributions
$P ( {\vec{\Theta}})$, 
we select independent weakly informed priors with
log-normal distributions for the three physics parameters, 
$\vec{\vartheta}=\{ c, k, x_0 \}$,
%$\vec{\vartheta}=\{ A_0, c, k, x_0 \}$,
normal distributions for the network parameters, 
$\vec{\theta}=\{ \mat{W}_k, \mat{b}_k \}$,
\beq
\begin{array}{@{\hspace*{0.0cm}}l@{\hspace*{0.0cm}}l}
  P ( {\vec{\Theta}})
= P ( c, k, x_0, \mat{W}_k, \mat{b}_k )
%= P ( A_0, c, k, x_0, \mat{W}_k, \mat{b}_k )
  \qquad \mbox{with} \qquad   
\begin{array}{ @{\hspace*{0.0cm}}l@{\hspace*{0.1cm}}
              c@{\hspace*{0.1cm}}
              c@{\hspace*{0.1cm}}
              l@{\hspace*{0.1cm}}l}                          
%\log &(A_0) &\sim& {\mathcal{N}}(\mu=\log(0.5), & \sigma=0.5)    \\
\log &(c)   &\sim& {\mathcal{N}}(\mu=\log(2.2), & \sigma=0.5)    \\
\log &(k)   &\sim& {\mathcal{N}}(\mu=\log(350), & \sigma=0.5)    \\
\log &(x_0) &\sim& {\mathcal{N}}(\mu=\log(0.56),& \sigma=0.5)    \\
&\mat{W}_k  &\sim& {\mathcal{N}}(\mu=0.0, & \sigma=2.0)    \\
&\mat{b}_k  &\sim& {\mathcal{N}}(\mu=0.0, & \sigma=2.0) \,.
\end{array}
\end{array}
\label{BPINNpriors}
\eeq
Using the case data from Figure \ref{fig01}, we approximately infer the posterior distributions from Bayes theorem (\ref{BNNposterior}) by employing a Hamiltonian Monte Carlo sampler implementation of Tensorflow-Probability \cite{tensorflow20}. For the Hamiltonian Monte Carlo sampler, we use 50 leapfrog steps with an initial time step of $\sca{d}t = 5 \cdot 10^{-4}$, and set the burn-in steps and total number of samples to 3000. We use the first 3000 samples to tune the sampler and the subsequent 3000 samples to estimate the set of parameters ${\vec{\Theta}}$.
%\begin{itemize}
%\item all settings were exactly the same as for the BNN
%\item for the ODE in the PINN we used the exact same priors for $c, \, k, \, x_0$ as for the Bayesian Inference model
%\end{itemize}
%%%%%%%%%%%%%%%%%%%%%%%%%%%%%%%%%%%%%%%%%%%%%%%%%%%%%%%%%%%%%%%%%%%
\subsection{Comparison of Bayesian Inference modeling}\label{BIEX}
\begin{figure}[h]
\centering
\includegraphics[width=0.8\linewidth]{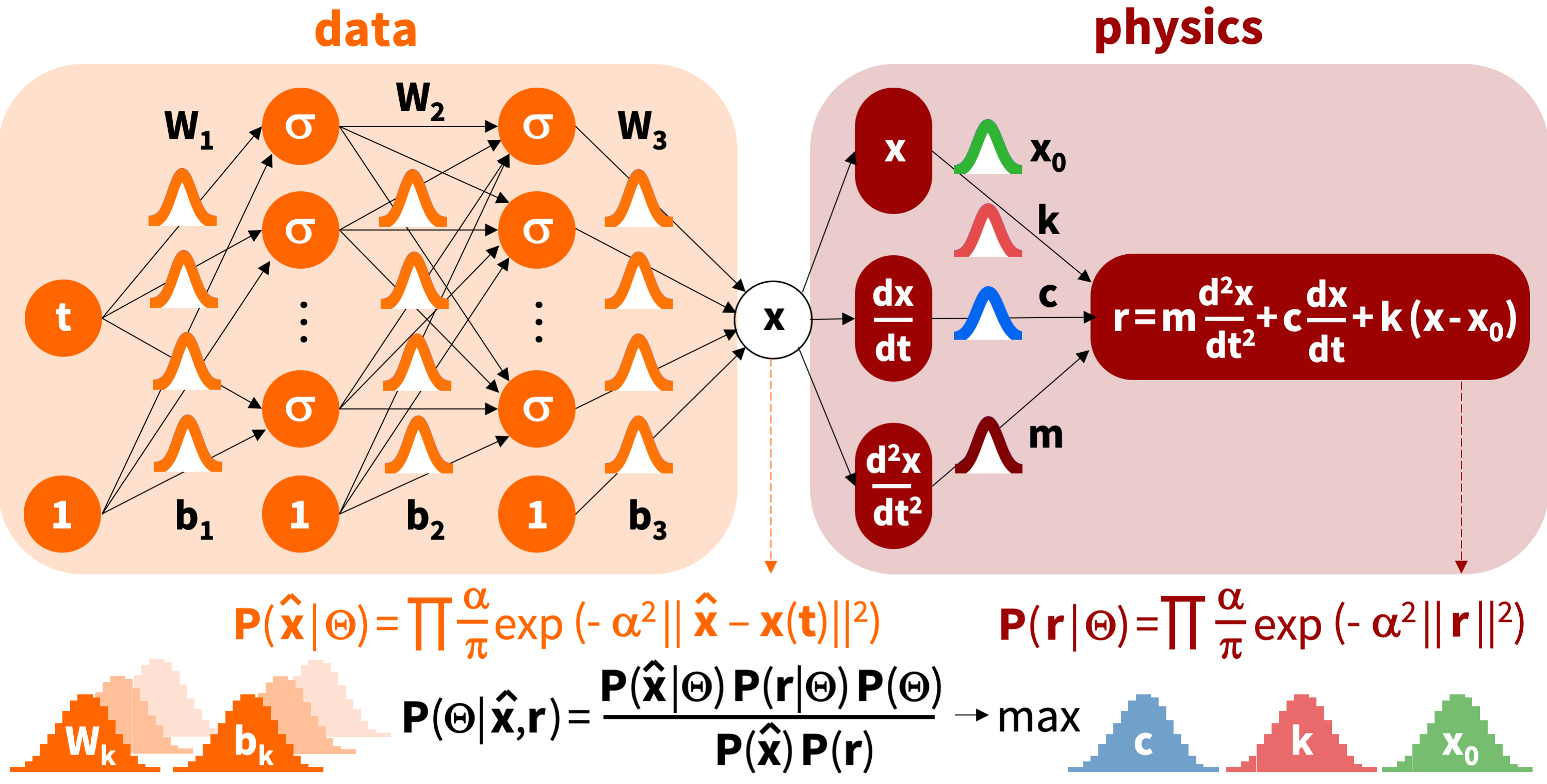}
\caption{{\bf{\sffamily{Bayesian Inference modeling.}}} 
Bayesian Inference maximizes a 
prior-weighted likelihood function 
$P(\vec{\Theta}|\hat{\mat{x}},\mat{r})\rightarrow\rm{max}$ 
that could consist of two terms, 
the likelihoods 
$P(\hat{\mat{x}}|\vec{\Theta})$ and $P({\mat{r}}|\vec{\Theta})$
of data $\hat{\mat{x}}$ and physics ${\mat{r}}$ 
for given parameters $\vec{\Theta}$,
to infer distributions 
of the network parameters 
$\vec{\theta} = \{ \mat{W}_k, \mat{b}_k \}$ and
physics parameters
$\vec{\vartheta} = \{ c, k, x_0 \}$.}
\label{figBI}
\end{figure}

\begin{figure}[h]
\centering
\includegraphics[width=0.8\linewidth]{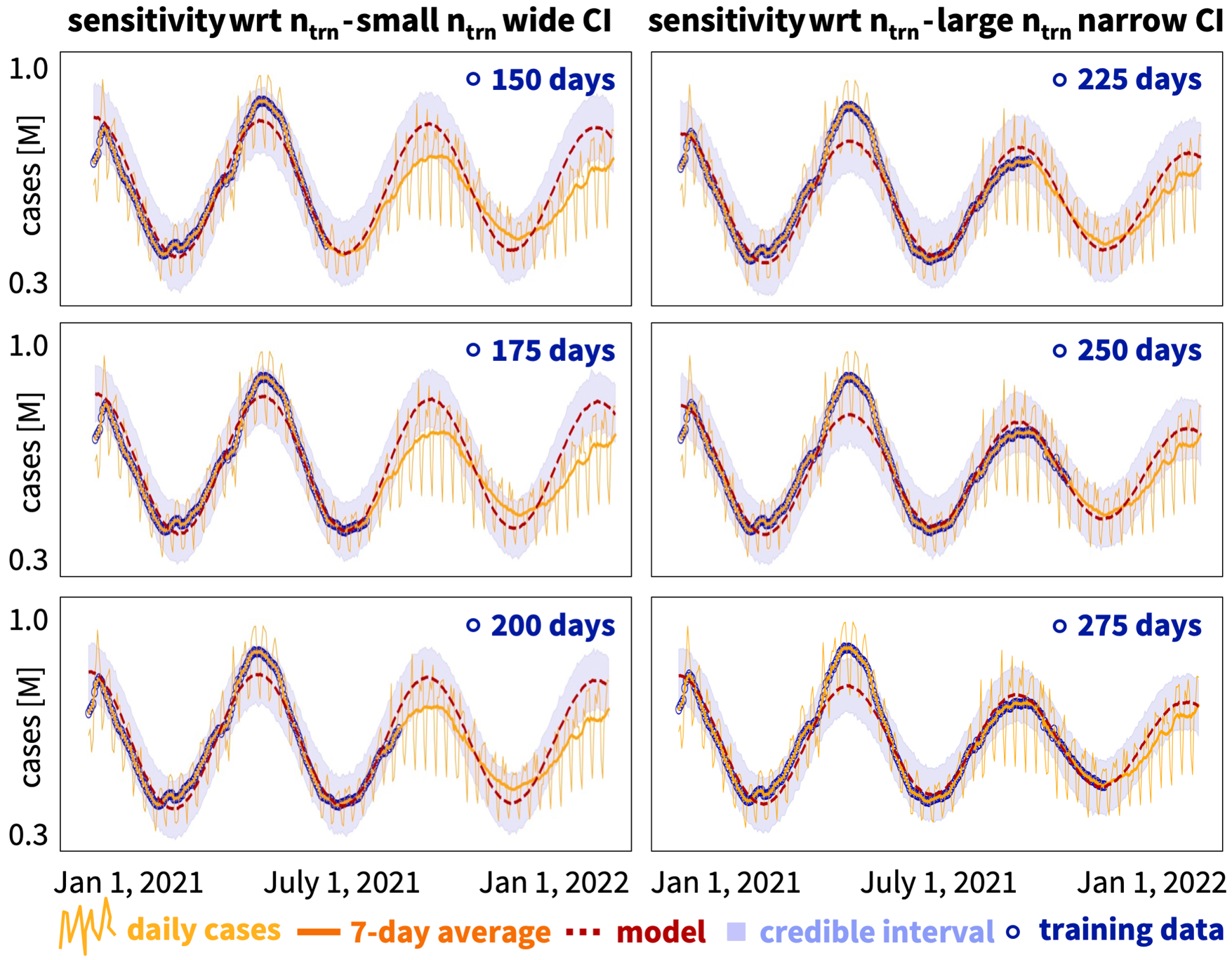}
\caption{{\bf{\sffamily{Bayesian Inference.
Sensitivity analysis for varying size of training data set.}}}
With increasing training data size of
150, 175, 200, 225, 250, and 275  days,
the error in the descriptive window increases, 
the error in the predictive  window decreases, and
the width of the credible interval decreases (top left to bottom right). 
Thin yellow lines indicate daily new case data, thick orange lines are their seven-day average, dashed red lines are the physics-based model, light blue areas are its credible intervals, blue dots are the training data.}
\label{fig07}
\end{figure}

%%%%%%%%%%%%%%%%%%%%%%%%%%%%%%%%%%%%%%%%%%%%%%%%%%%%%%%%%%%%%%%%%%%
\noindent
Figure \ref{figBI} illustrates the features the Bayesian Inference models of this section. 
The classical Bayesian Inference from Section \ref{BI}
maximizes a prior-weighted likelihood, 
$P(\hat{\mat{x}}|\vec{\vartheta})$,
in terms of the error between data and model
$|| \, \hat{x} - {x}(t) \, ||$ 
to infer distributions of the physics parameters $\vec{\vartheta}$.
The Bayesian Neural Networks from Section \ref{BNN}
maximize a prior-weighted likelihood, 
$P(\hat{\mat{x}}|\vec{\theta})$,
in terms of the error between data and model
$|| \, \hat{x} - {x}(t) \, ||$
to infer distributions of the network parameters $\vec{\theta}$.
The Bayesian Physics Informed Neural Networks from Section \ref{BPINN}
maximize a prior-weighted likelihood, 
$P(\hat{\mat{x}},\mat{r}|\vec{\Theta})$,
in terms of the error between data and model
$|| \, \hat{x} - {x}(t) \, ||$, shown on the left, 
and the error of the physics 
$|| \, r \, ||$, shown on the right,
to infer the distributions of both
network and physics 
parameters $\vec{\theta}$ and $\vec{\vartheta}$.
%%%%%%%%%%%%%%%%%%%%%%%%%%%%%%%%%%%%%%%%%%%%%%%%%%%%%%%%%%%%%%%%%%%%%%%%
\\[4.pt]
%%%%%%%%%%%%%%%%%%%%%%%%%%%%%%%%%%%%%%%%%%%%%%%%%%%%%%%%%%%%%%%%%%%%%%%%
\noindent
Figure \ref{fig07} illustrates the results of the classical Bayesian Inference for a varying size of the training data set.
The thin yellow lines indicate daily new case data, the thick orange lines are their seven-day average, and the blue dots are the training data varying between 
150, 175, 200, 225, 250, and 275 days, from top left to bottom right. 
The dashed red lines represent the physics-based model and the light blue areas are its credible intervals.
%%%%%%%%%%%%%%%%%%%%%%%%%%%%%%%%%%%%%%%%%%%%%%%%%%%%%%%%%%%%%%%%%%%
%%%%%%%%%%%%%%%%%%%%%%%%%%%%%%%%%%%%%%%%%%%%%%%%%%%%%%%%%%%%%%%%%%%
First and foremost, we note that all six panels display a reasonable agreement between data and model, both inside the training window and beyond. 
With an increasing size of training data,
from top left to bottom right,
the error in the descriptive window increases and
the error in the predictive window decreases.
For example, 
the first peak of 905,378 cases on day 118, April 29, 2021,
is best approximated for the smallest training data set in the top left,
% first  valley of 281,223 on day 46, February 16, 2021,
% second valley of 296,808 on day 172, June 22, 2021,
while the second peak of 819,336 cases on day 221, August 10, 2021,
is best approximated for the largest training data set in the bottom right. 
In agreement with our expectation, 
the width of the credible interval,
the light blue area,
decreases with an increase in training data:  
The more data we use to train the model, the better its fit. 
%%%%%%%%%%%%%%%%%%%%%%%%%%%%%%%%%%%%%%%%%%%%%%%%%%%%%%%%%%%%%%%%%%%
\begin{figure}
\centering
\includegraphics[width=1.0\linewidth]{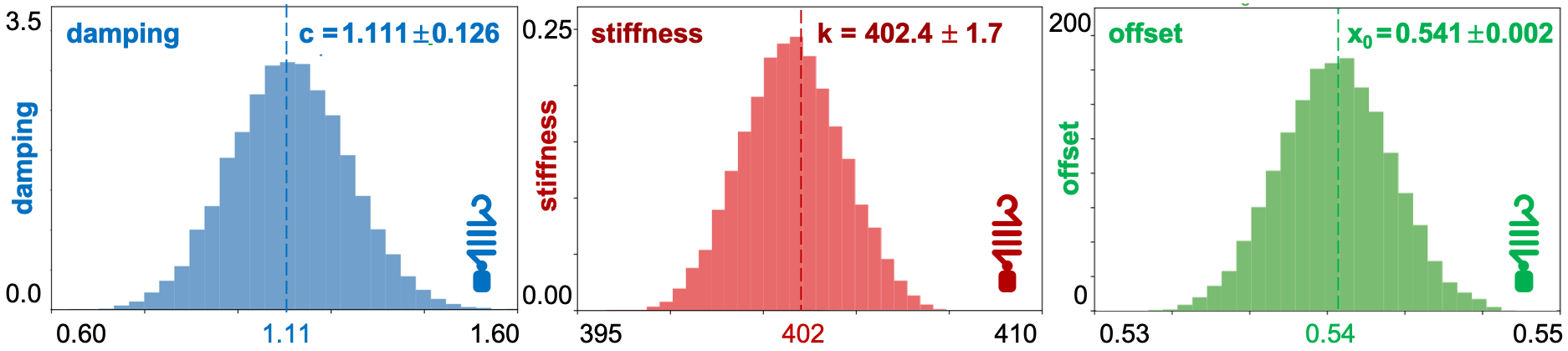}
\caption{{\bf{\sffamily{Bayesian Inference. 
Posterior distributions of physics parameters
damping $\vec{c}$, stiffness $\vec{k}$, and offset $\vec{x_0}$.}}} 
Inferred
viscous damping coefficient of $c=1.111 \pm 0.126$, 
stiffness of $k=402.4 \pm 1.7$, and 
offset of $x_0=0.541 \pm 0.002$.
This corresponds to 
a damping of $\delta=0.556$,
a frequency of $\omega = 20.06$,
and a period of $T=0.313$ years.
Simulation with training window of $n_{\rm{trn}}$ = 225 days.}
\label{fig08}
\end{figure}\\[6.pt]
%%%%%%%%%%%%%%%%%%%%%%%%%%%%%%%%%%%%%%%%%%%%%%%%%%%%%%%%%%%%%%%%%%%
Figure \ref{fig08} illustrates the posterior distributions of the physics parameters $\vec{\vartheta}$, the damping $c$, stiffness $k$, and offset $x_0$. 
The Bayesian Inferences yields   
a viscous damping coefficient of $c=1.111 \pm 0.126$, 
a stiffness of $k=402.4 \pm 1.7$, and 
an offset of $x_0=0.541 \pm 0.002$.
This corresponds to 
a damping of $\delta=0.556$,
a frequency of $\omega = 20.06$,
and a period of $T=0.313$ years.
These values compare well to the learned parameters of the Physics Informed Neural Network in Figure \ref{fig06}, with 
a viscous damping coefficient of $c=1.251$, 
a stiffness of $k=374.6$, and 
an offset of $x_0=0.558$. 
However, the Bayesian Inference not only inferred the means of the parameters, but also their posterior distributions, where narrow distributions indicate a higher confidence in the mean values. 
%%%%%%%%%%%%%%%%%%%%%%%%%%%%%%%%%%%%%%%%%%%%%%%%%%%%%%%%%%%%%%%%%%%
\begin{figure}
\centering
\includegraphics[width=1.0\linewidth]{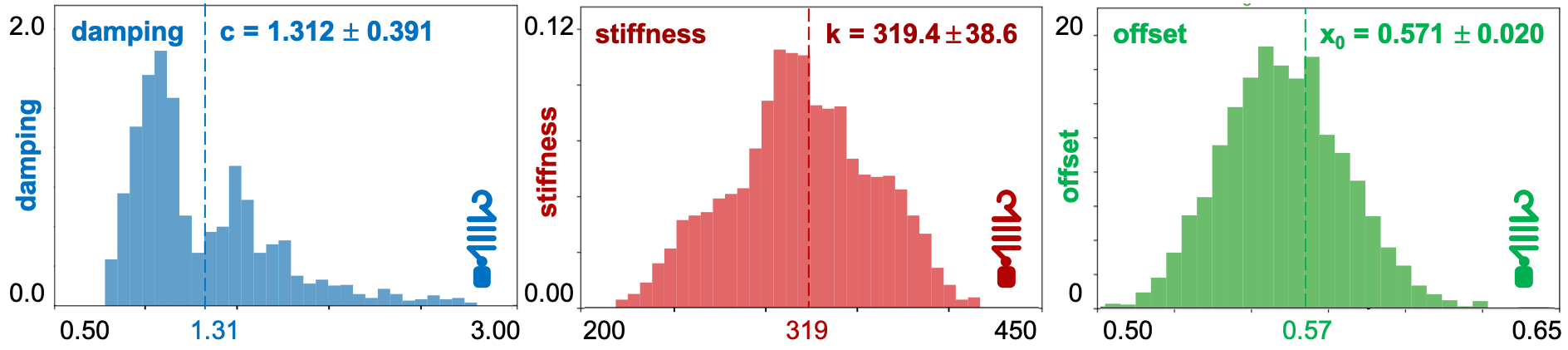}
\caption{{\bf{\sffamily{Bayesian Physics Informed Neural Network. 
Posterior distributions of physics parameters
damping $\vec{c}$, stiffness $\vec{k}$, and offset $\vec{x_0}$.}}} 
Inferred
viscous damping coefficient of $c=1.312 \pm 0.391$, 
stiffness of $k=319.4 \pm 38.6$, and 
offset of $x_0=0.571 \pm 0.020$.
This corresponds to 
a damping of $\delta=0.65$,
a frequency of $\omega = 17.87$,
and a period of $T=0.352$ years.
Simulation with training window of $n_{\rm{trn}}$ = 225 days.}
\label{fig14}
\end{figure}\\[6.pt]
%%%%%%%%%%%%%%%%%%%%%%%%%%%%%%%%%%%%%%%%%%%%%%%%%%%%%%%%%%%%%%%%%%%
Figure \ref{fig14} illustrates the posterior distributions of the physics parameters $\vec{\vartheta}$, the damping $c$, stiffness $k$, and offset $x_0$.
The Bayesian Physics Informed Neural Network yields   
a viscous damping coefficient of $c=1.312 \pm 0.391$, 
a stiffness of $k=319.4 \pm 38.6$, and 
an offset of $x_0=0.571 \pm 0.020$.
This corresponds to 
a damping of $\delta=0.65$,
a frequency of $\omega = 17.87$,
and a period of $T=0.352$ years.
Compared to the classical Bayesian Inference in Figure \ref{fig08}, 
the damping is 18\% larger, 
the stiffness is 21\% smaller, and
the offset is 6\% larger.  
Notably, all three parameters of the Bayesian Physics Informed Neural Network show wider standard deviations than in the classical Bayesian Inference, the damping by a factor 3, stiffness by a factor 23, and offset by a factor 10. 
These observations suggests that, for the chosen training window of $n_{\rm{trn}}$ = 225 days, the Bayesian Physics Informed Neural Network is not only more expensive than the Bayesian Inference, but also less robust.  
%%%%%%%%%%%%%%%%%%%%%%%%%%%%%%%%%%%%%%%%%%%%%%%%%%%%%%%%%%%%%%%%%%%
\section{Discussion}\label{discussion}
%%%%%%%%%%%%%%%%%%%%%%%%%%%%%%%%%%%%%%%%%%%%%%%%%%%%%%%%%%%%%%%%%%%
\noindent
Understanding the behavior of nonlinear dynamical systems remains a challenging task. Machine learning has emerged as a powerful technology to provide insight into observational data. 
%%%
A vast variety of learning machines have been developed throughout the past decade, but to the unexperienced user, selecting the right model  for a specific task is an overwhelming endeavor.  
%%%
This is amplified by the fact that tutorials or publications typically introduce new methods using artificially generated synthetic data for which the solution is already known.   
%%%
The objective of this study was to compare different families of models for real-world data, the reported number of COVID-19 cases throughout the year of 2021, using  
a simple dynamical model, a damped harmonic oscillator.
%%%
We reviewed the underlying model equations, parameters, and characteristic results of two families of models, Neural Networks and Bayesian Inference, each with three members of increasing complexity. %\\[6.pt]
%%%%%%%%%%%%%%%%%%%%%%%%%%%%%%%%%%%%%%%%%%%%%%%%%%%%%%%%%%%%%%%%%%%
\begin{figure}[h]
\centering
\includegraphics[width=0.5\linewidth]{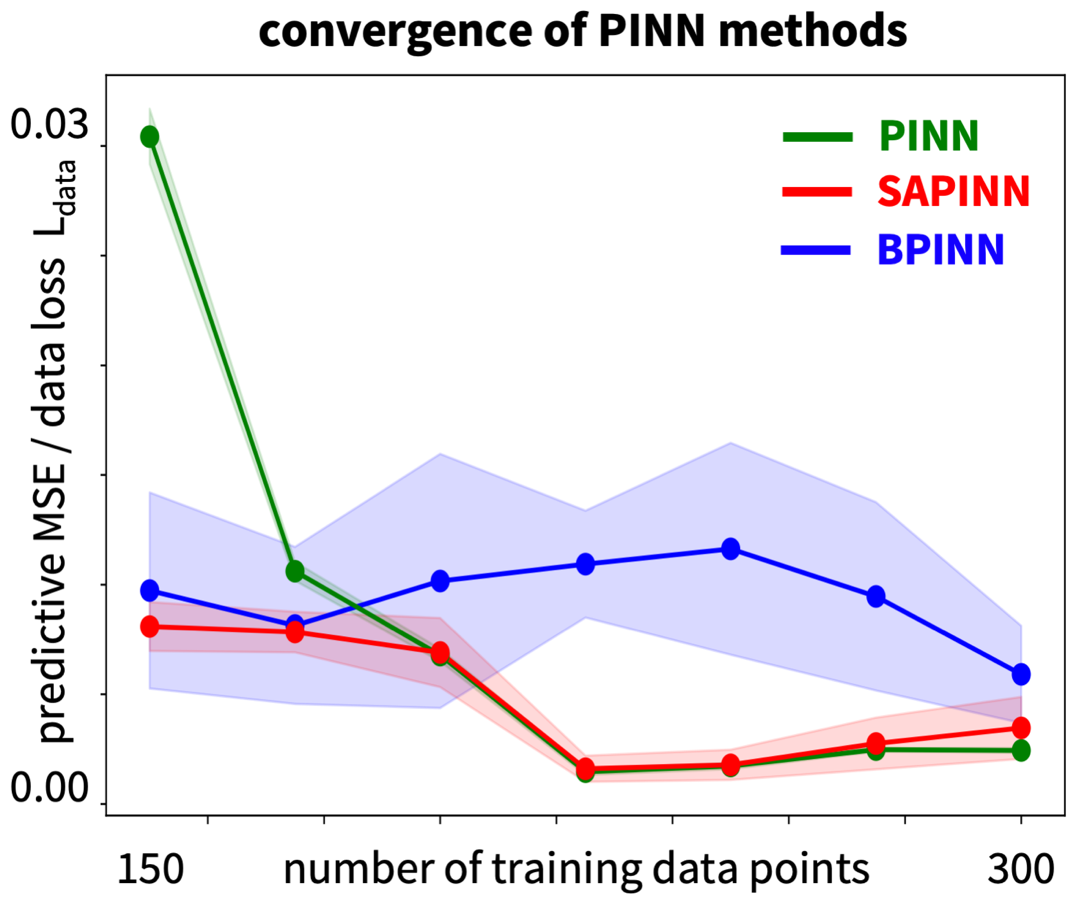}
\caption{{\bf{\sffamily{Comparison of Physics Informed Neural Network models.}}} 
Convergence with increasing size of training data set.
Predictive data loss, 
$L_{\rm{data}}$, 
as mean squared error in the predictive window 
from $t^{\rm{n}_{\rm{trn}}}$ to $t^{\rm{n}_{\rm{smp}}}$,
for
Physics Informed Neural Network,
Self Adaptive Physics Informed Neural Network, and 
Bayesian Physics Informed Neural Network.
For all three methods, the predictive data loss decreases with an increasing number of training data points.
The Physics Informed Neural Network performs poorly initially, but converges with increasing training set size.
The Self Adaptive Physics Informed Neural Network performs well, even for small training sets.
Both methods perform similarly with increasing training set size. 
The Bayesian Physics Informed Neural Network performs reasonably well for small data sets but only improves marginally with increasing training set size.
Training data 
$\rm{n}_{\rm{trn}}$= [150, 175, 200, 225, 250, 275, 300].
}
\label{fig11}
\end{figure}
%%%%%%%%%%%%%%%%%%%%%%%%%%%%%%%%%%%%%%%%%%%%%%%%%%%%%%%%%%%%%%%%%%%
%%%%%%%%%%%%%%%%%%%%%%%%%%%%%%%%%%%%%%%%%%%%%%%%%%%%%%%%%%%%%%%%%%%
\begin{figure}[h]
\centering
\includegraphics[width=0.8\linewidth]{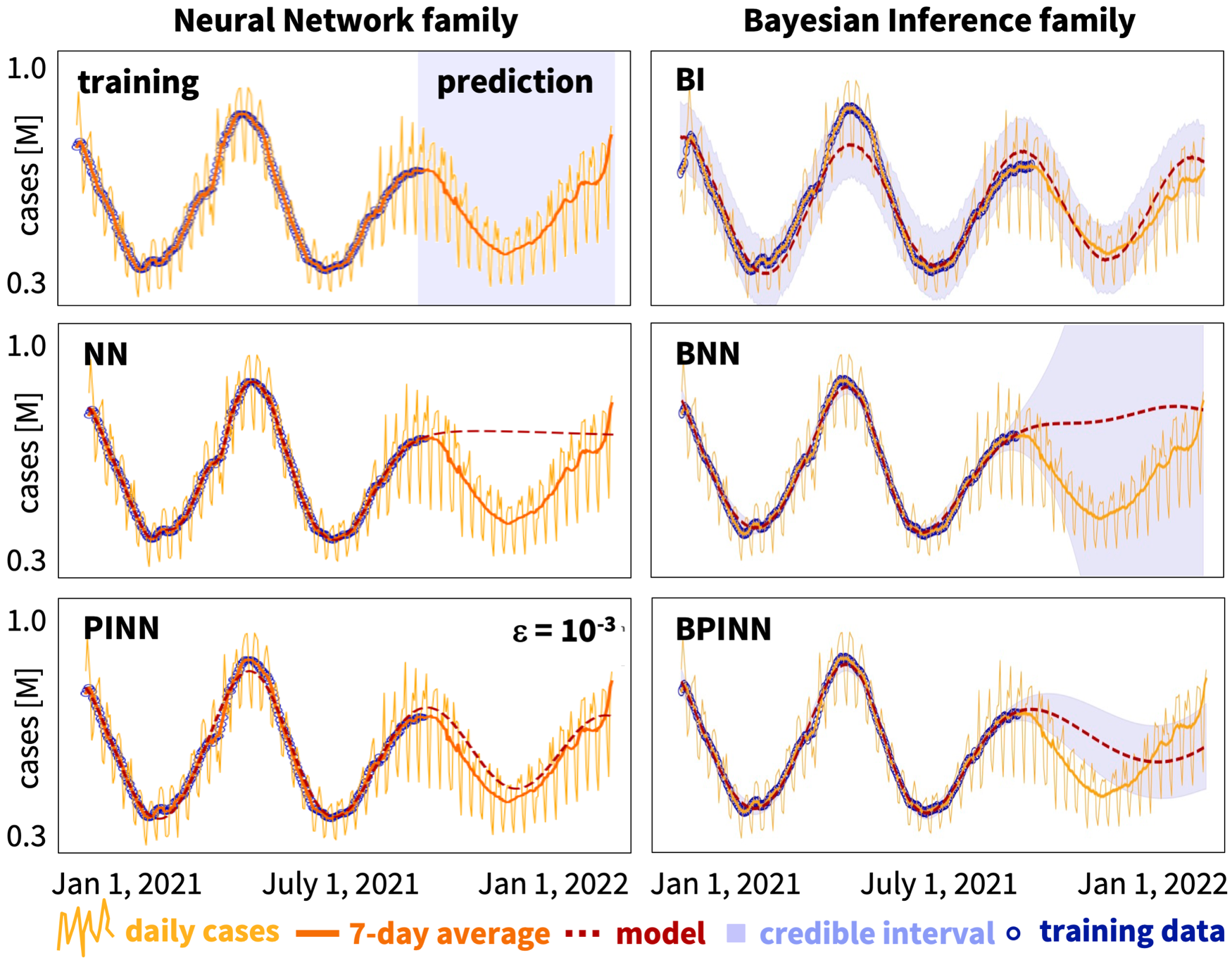}
\caption{{\bf{\sffamily{Comparison of Neural Network and Bayesian Inference models.}}}
Neural Networks approximate the training data well, but fail to predict the behavior outside the training window (top left).
Physics Informed Neural Networks approximate both the training data and the behavior outside the training window, but are sensitive to the weighting coefficient (middle left).
Self Adaptive Physics Informed Neural Networks approximate both the training data and the behavior outside the training window, and simultaneously 
learn the weighting term $\varepsilon(t)$ (bottom left).
% ...
Bayesian Inference not only approximates both, the training data and the behavior outside the training window, but also provides credible intervals of the model (top right).
Bayesian Neural Networks approximate the training data well, but fail to predict the behavior outside the training window where they generate a wide credible intervals (middle right). 
Bayesian Physics Informed Neural Networks approximate the training data well, but fail to predict the behavior outside the training window (bottom right).
% ...
Thin yellow lines indicate daily new case data, thick orange lines are their seven-day average, dashed red lines are the model, light blue areas are its credible intervals, blue dots are the training data.}
\label{fig10}
\end{figure}
%%%%%%%%%%%%%%%%%%%%%%%%%%%%%%%%%%%%%%%%%%%%%%%%%%%%%%%%%%%%%%%%%%%
%%%%%%%%%%%%%%%%%%%%%%%%%%%%%%%%%%%%%%%%%%%%%%%%%%%%%%%%%%%%%%%%%%%
\begin{figure}
\centering
\includegraphics[width=0.8\linewidth]{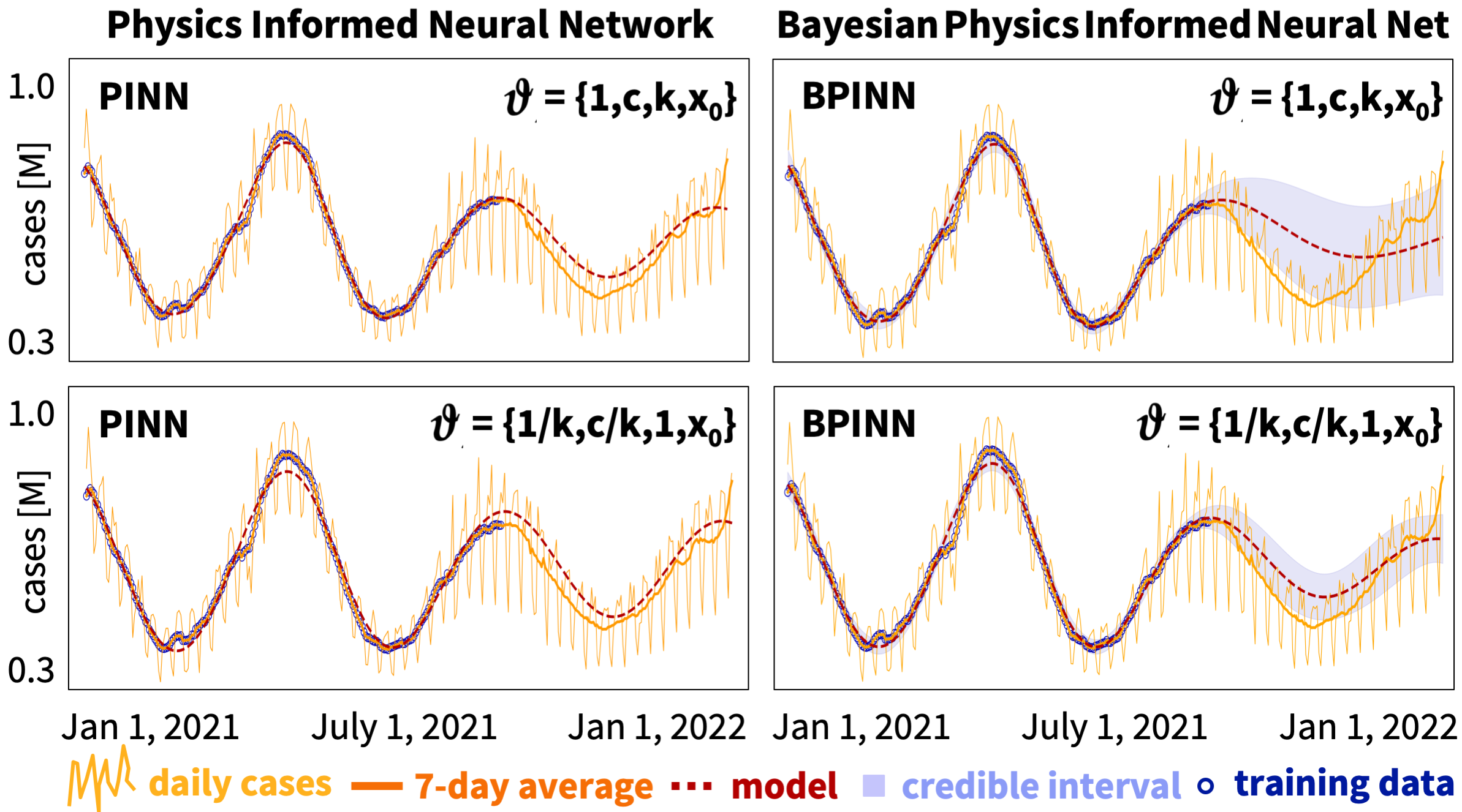}
\caption{{\bf{\sffamily{Comparison of classical and Bayesian
Physics Informed Neural Network models for different scaling.}}} 
Both types of models are sensitive to the scaling of the physics equation,
$\ddot{x}   + c \, \dot{x}   + k (\,x - x_0\,)$ (top row) and
$\ddot{x}/k + c \, \dot{x}/k + (\,x - x_0\,)$ (bottom row).
Appropriate scaling improves model performance:
The fit between the red dashed lines of the model $x(t)$ and the thick orange lines of the data $\hat{x}$ increases and the light blue credible intervals decrease.
Thin yellow lines indicate daily new case data, thick orange lines are their seven-day average, dashed red lines are the model, light blue areas are its credible intervals, blue dots are the training data.}
\label{fig_conditioning}
\end{figure}
%%%%%%%%%%%%%%%%%%%%%%%%%%%%%%%%%%%%%%%%%%%%%%%%%%%%%%%%%%%%%%%%%%%
%%%%%%%%%%%%%%%%%%%%%%%%%%%%%%%%%%%%%%%%%%%%%%%%%%%%%%%%%%%%%%%%%%%
\begin{table}[h]
\centering
\caption{{\bf{\sffamily{Comparison of Neural Network and Bayesian Inference models.}}} Advantages and disadvantages of Neural Networks, Physics Informed Neural Networks, Self Adaptive Physics Informed Neural Networks, Bayesian Inference, Bayesian Neural Networks, and Bayesian Physics Informed Neural Networks in terms of parameters, data fit, predictive potential, and robustness.}
\label{tab01}
\vspace*{0.2cm} 
\hspace*{-1.5cm}
{\small{
\begin{tabular}{|@{\hspace*{0.3cm}}l@{\hspace*{0.1cm}}|
                |@{\hspace*{0.2cm}}l@{\hspace*{0.2cm}}|} \hline                
%%%%%%%%%%%%%%%%%%%%%%%%%%%%%%%%%%%%%%%%%%%%%%%%%%%%%%%%%%%%%%%%%%%
  {\bf{\sffamily{Neural Network family}}} 
& {\bf{\sffamily{Bayesian Inference family}}} \\ \hline \hline
%%%%%%%%%%%%%%%%%%%%%%%%%%%%%%%%%%%%%%%%%%%%%%%%%%%%%%%%%%%%%%%%%%%
  {\bf{\sffamily{NN}}}
& {\bf{\sffamily{BI}}}         \\ \hline 
%%%%%%%%%%%%%%%%%%%%%%%%%%%%%%%%%%%%%%%%%%%%%%%%%%%%%%%%%%%%%%%%%%%
% simple and straightforward      & simple and straightforward \\
  fits network without physics, no credible intervals     
& fits physics model, provides credible intervals \\
  learns $\vec{\theta}    = \{ \mat{W}_k, \mat{b}_k \}$      
& learns distributions of $\vec{\vartheta} = \{ c, k, x_0 \}$\\
  here 32+32$\times$32+32+32+32+1=1153 unknowns
& here 3$\times$2=6 unknowns \\
  good fit of training data,   poor predictive potential      
& decent fit of training data, good predictive potential \\
  simple, robust, and straightforward            
& simple, robust, and straightforward  \\ \hline  \hline 
%%%%%%%%%%%%%%%%%%%%%%%%%%%%%%%%%%%%%%%%%%%%%%%%%%%%%%%%%%%%%%%%%%%
  {\bf{\sffamily{PINN}}}
& {\bf{\sffamily{BNN}}}         \\ \hline 
%%%%%%%%%%%%%%%%%%%%%%%%%%%%%%%%%%%%%%%%%%%%%%%%%%%%%%%%%%%%%%%%%%%
  fits network and physics model, no credible intervals   
& fits network without physics,provides\,credible\,intervals \\
  learns 
  $\vec{\theta}    = \{ \mat{W}_k, \mat{b}_k \}$ and 
  $\vec{\vartheta} = \{ c, k, x_0 \}$ 
& learns distributions of 
  $\vec{\theta}    = \{ \mat{W}_k, \mat{b}_k \}$\\
  here 32+32$\times$32+32+32+32+1+3=1156 unknowns
& here (32+32$\times$32+32+32+32+1)$\times$2=2306 unknowns \\
  good fit of training data, good predictive potential      
& good fit of training data, poor predictive potential \\
  sensitive to choice of weighting coefficient $\varepsilon$           
& credible intervals provide insight into reliable regimes \\ \hline  \hline 
%%%%%%%%%%%%%%%%%%%%%%%%%%%%%%%%%%%%%%%%%%%%%%%%%%%%%%%%%%%%%%%%%%%
  {\bf{\sffamily{SAPINN}}}
& {\bf{\sffamily{BPINN}}}         \\ \hline 
%%%%%%%%%%%%%%%%%%%%%%%%%%%%%%%%%%%%%%%%%%%%%%%%%%%%%%%%%%%%%%%%%%%
  adaptively fits network and physics,no\,credible\,intervals  
& fits network and physics, provides credible intervals \\
  learns 
  $\vec{\theta}    = \{ \mat{W}_k, \mat{b}_k \}$ and 
  $\vec{\vartheta} = \{ c, k, x_0 \}$ and $\varepsilon (t_{\rm{data}})$
& learns distributions of 
  $\vec{\theta}    = \{ \mat{W}_k, \mat{b}_k \}$ and
  $\vec{\vartheta} = \{ c, k, x_0 \}$\\
  here 32+32$\times$32+32+32+32+1+3+1=1157 unknowns
& here (32+32$\times$32+32+32+32+1+3)$\times$2=2312 unknowns \\
  good fit of training data, good predictive potential      
& good fit of training data, moderate predictive potential \\
  robust even for small training data set         
& most expensive method, requires large training data set \\ \hline
%%%%%%%%%%%%%%%%%%%%%%%%%%%%%%%%%%%%%%%%%%%%%%%%%%%%%%%%%%%%%%%%%%%
\end{tabular}
}}
\end{table}
%%%%%%%%%%%%%%%%%%%%%%%%%%%%%%%%%%%%%%%%%%%%%%%%%%%%%%%%%%%%%%%%%%%
%%%%%%%%%%%%%%%%%%%%%%%%%%%%%%%%%%%%%%%%%%%%%%%%%%%%%%%%%%%%%%%%%%%
\noindent
Both Neural Networks and Bayesian Inference
perform equally well on the raw daily case data with noise, reporting uncertainties, and weekday-weekend fluctuations and on their smooth seven-day average, as 
Figure \ref{fig03} confirms. Our learned damping, stiffness, and offset of
$c=1.251$, $k=374.6$, and $x_0=0.558$ 
for the Physics Informed Neural Network in Section \ref{PINN}, Figure \ref{fig06},
agree well with the inferred means of
$c=1.111$, $k=402.4$, $x_0=0.541$
for the Bayesian Inference in Section \ref{BI}, Figure \ref{fig08},
and with the combination of both methods of
$c=1.312$, $k=319.4$, $x_0=0.571$
for the Bayesian Physics Informed Neural Network in Section \ref{BPINN}, Figure \ref{fig14}.
Our study confirms our general intuition that the quality of the fit increases with increasing size of the training data set. For the classical Bayesian Inference in Section \ref{BI}, Figure \ref{fig07} shows that the credible intervals become progressively narrower as the size of the training data increases. 
Figure  \ref{fig11} compares the convergence of the three Physics Informed Neural Network models of our study, 
the plain Physics Informed Neural Network from Section \ref{PINN}
the Self Adaptive Physics Informed Neural Network from Section \ref{SAPINN}, and 
the Bayesian Physics Informed Neural Network from Section \ref{BPINN}.
Clearly, for all three methods, the prediction error decreases with increasing size of the training data set.
We close our study with a direct side-by-side comparison of all six methods from Sections \ref{NNmodeling} and \ref{BImodeling} in Figure \ref{fig10}.
Figure \ref{fig_conditioning} provides a final recommendation to improve the performance of the Bayesian Physics Informed Neural Networks by scaling the physics equation.
Table \ref{tab01} summarizes the results of Figures \ref{fig10} and \ref{fig_conditioning} and discusses the advantages and disadvantages of each method.  \\[6.pt]
%%%%%%%%%%%%%%%%%%%%%%%%%%%%%%%%%%%%%%%%%%%%%%%%%%%%%%%%%%%%%%%%%%%
\noindent{\bf{\sffamily{Neural Networks are a simple and robust tool to fit training data, but have a poor predictive potential.}}} 
%%%%%%%%%%%%%%%%%%%%%%%%%%%%%%%%%%%%%%%%%%%%%%%%%%%%%%%%%%%%%%%%%%%
Classical Neural Networks have remarkable power to fit a network model to data without any underlying physics. By minimizing a loss function that characterizes the error between model and data, the neural network learns the network parameters 
$\vec{\theta} = \{ \mat{W}_k, \mat{b}_k \}$, the network weights and biases. For our example of a feed-forward network with one input, two hidden layers with 32 nodes each, and one output, 
it has 32+32$\times$32+32=1088 weights and 32+32+1=65 biases,
resulting in a total of 1153 unknowns.
Figure \ref{fig10}, top left, suggests that even in the absence of any prior physical knowledge, the classical Neural Network provides an excellent approximation of the training data. However, it fails to predict the behavior outside the training window, where it simply continues the linear horizontal trend from the last set of training data points. This becomes particularly clear when the underlying problem is non-monotonic, or, like in our case, even oscillatory. Since classical Neural Networks provide no credible intervals, we have no way of knowing, how poor the predictive potential of the network really is. \\[6.pt]
%\clearpage
%%%%%%%%%%%%%%%%%%%%%%%%%%%%%%%%%%%%%%%%%%%%%%%%%%%%%%%%%%%%%%%%%%%
\noindent{\bf{\sffamily{Physics Informed Neural Networks integrate data and physics and have a good predictive potential.}}} 
%%%%%%%%%%%%%%%%%%%%%%%%%%%%%%%%%%%%%%%%%%%%%%%%%%%%%%%%%%%%%%%%%%%
Physics Informed Neural Networks fit a network model to data, and, at the same time, to a physics-based model. 
If we know the underlying physics, they are an effective tool to constrain a deep learning method to a lower-dimensional manifold and create models that can be trained effectively with a small amount of data.
Physics Informed Neural Networks enforce physics via soft penalty constraints. 
By minimizing a loss function that characterizes the error between model and data and the error in satisfying the physics, they simultaneously learn the network parameters 
$\vec{\theta} = \{ \mat{W}_k, \mat{b}_k \}$ and the physics parameters
$\vec{\vartheta} = \{ c, k, x_0 \}$.
For our example,
the network has 1088 weights and 65 biases,
and the physics model has a damping, stiffness, and offset parameter,
resulting in a total of 1156 unknowns.
Figure \ref{fig10}, middle left, suggests that the Physics Informed Neural Network
approximates both the training data and the behavior outside the training window reasonably well and is equally capable of both interpolation and extrapolation.
From Figure \ref{fig11} we conclude that Physics Informed Neural Networks require a relatively large set of training data. For our example, they perform moderately initially, but converge well with increasing training set size.
However, from Figures \ref{fig04} and~\ref{fig05}, we conclude that Physics Informed Neural Networks are sensitive to the weighting coefficient $\varepsilon$ that can bias the solution to emphasize on either data or physics.
From Figure \ref{fig_conditioning}, we add that they are also sensitive to a scaling of the physics parameters $\vec{\vartheta}$, which, if done appropriately, can improve the performance of the model.
An inherent limitation of plain Physics Informed Neural Networks is that they are not equipped with built-in uncertainty quantification which may restrict their applications, especially in situations with noisy data.\\[6.pt]
%%%%%%%%%%%%%%%%%%%%%%%%%%%%%%%%%%%%%%%%%%%%%%%%%%%%%%%%%%%%%%%%%%%
\noindent{\bf{\sffamily{Self Adaptive Physics Informed Neural Networks provide adaptive and robust fits of data and physics.}}} 
%%%%%%%%%%%%%%%%%%%%%%%%%%%%%%%%%%%%%%%%%%%%%%%%%%%%%%%%%%%%%%%%%%%
Self Adaptive Physics Informed Neural Networks adaptively fit a network model to data, and, at the same time, to a physics-based model. 
They inherit all the advantages of Physics Informed Neural Networks and address the limitation of bias between data and physics by introducing the weighting coefficient as independent time-varying unknown.
This allows them to perform well, even in regions with steep gradients, while using a smaller number of training epochs. % \cite{mcclenny20}. 
Self Adaptive Physics Informed Neural Networks minimize a loss function that characterizes the error between model and data and between model and physics, and learn the network parameters 
$\vec{\theta} = \{ \mat{W}_k, \mat{b}_k \}$, the physics parameters
$\vec{\vartheta} = \{ c, k, x_0 \}$, and the weighting term 
$\varepsilon(t)$ between network and physics. 
For our example,
the network model has 1088 weights and 65 biases,
the physics model has a damping, stiffness, and offset,
and the self adaptive model has a weighting coefficient,
resulting in a total of 1157 unknowns.
Figure \ref{fig10}, bottom left, confirms that the Self Adaptive Physics Informed Neural Network approximates both the training data and the behavior outside the training window, similar to the regular Physics Informed Neural Network.
With only one additional time-varying parameter, the self-adaptively learned weighting term, it performs well even for small training data sets, 
as Figure \ref{fig11} confirms. 
One caveat is that Self Adaptive Physics Informed Neural Networks involve training with complex loss functions that consist of multiple terms and result in a highly non-convex optimization problem.
During training, these terms compete with one another which implies that the training process may not be robust and stable, and does not always converge to a global minimum.
Self Adaptive Physics Informed Neural Networks perform best when the equations are well scaled and pre-conditioned and their parameters all lie within the same range of magnitude. 
While we have not explicitly shown the effect of different pre-conditioning and scaling techniques here, we have observed that the method is sensitive to scaling and could fail to converge when scaled inappropriately.\\[6.pt]
%%%%%%%%%%%%%%%%%%%%%%%%%%%%%%%%%%%%%%%%%%%%%%%%%%%%%%%%%%%%%%%%%%%
\noindent{\bf{\sffamily{Bayesian Inference fits a physics based model to data and provides credible intervals.}}} 
%%%%%%%%%%%%%%%%%%%%%%%%%%%%%%%%%%%%%%%%%%%%%%%%%%%%%%%%%%%%%%%%%%%
Classical Bayesian Inference is a simple, robust, and stable method that fits a physics based model to data. 
This implies that we know the underlying physics. 
In our example, we do not know the exact physics; however, we hypothesized that the data, the daily new cases of COVID-19 worldwide, follow the dynamics of a damped harmonic oscillator.
Bayesian Inference then uses Bayes' theorem to infer the distribution of a set of model parameters,  
$\vec{\vartheta} = \{ c,k,x_0 \}$,
that best explain the data. 
For our example, we have assumed log normal distributions for each parameter, with individual means $\mu$ and standard deviations $\sigma$, introducing a total of $3 \times 2=6$ unknowns.  
Figure \ref{fig10}, top right, demonstrates that, provided the physics-based model is reasonable, the Bayesian Inference approximates the behavior well throughout the entire time window. Instead of only inferring a point value for each parameter, Bayesian Inference infers parameter distributions and credible intervals that provide valuable insight into the goodness of the fit as we have seen in the sensitivity analysis for varying training data set sizes in Figure \ref{fig07}: The narrower the credible interval, the better the fit.\\[6.pt]
%%%%%%%%%%%%%%%%%%%%%%%%%%%%%%%%%%%%%%%%%%%%%%%%%%%%%%%%%%%%%%%%%%%
\noindent{\bf{\sffamily{Bayesian Neural Networks fit a network model to data and provide credible intervals.}}} 
%%%%%%%%%%%%%%%%%%%%%%%%%%%%%%%%%%%%%%%%%%%%%%%%%%%%%%%%%%%%%%%%%%%
Bayesian Neural Networks fit a network model to data without any underlying physics. 
They use Bayes' theorem to infer the distribution of a set of network parameters,  
$\vec{\theta}    = \{ \mat{W}_k, \mat{b}_k \}$ 
that best explain the data. 
For our example, a feed-forward network with one input, two hidden layers with 32 nodes each, and one output, 
the network has 1088 weights and 65 biases.
We have assumed log-normal distributions for each parameter, with individual means $\mu$ and standard deviations $\sigma$, introducing a total of $1053 \times 2 = 2306$ unknowns.  
Figure \ref{fig10}, middle right, shows that the Bayesian Neural Network shares the features of classical Neural Networks and Bayesian Inference. It approximates the training data well, but fails to predict the behavior outside the training window. 
The wide credible intervals highlight the poor predictive potential of the model and confirm that any prediction a few days beyond the training window generates unreliable results. \\[6.pt]
%%%%%%%%%%%%%%%%%%%%%%%%%%%%%%%%%%%%%%%%%%%%%%%%%%%%%%%%%%%%%%%%%%%
\noindent{\bf{\sffamily{Bayesian\,Physics\,Informed\,Neural\,Networks\,provide a good fit and prediction,\,but\,are\,sensitive\,to\,scaling.}}} 
%%%%%%%%%%%%%%%%%%%%%%%%%%%%%%%%%%%%%%%%%%%%%%%%%%%%%%%%%%%%%%%%%%%
Bayesian Physics Informed Neural Networks fit both network and physics, and, in addition, provide credible intervals that provide insight into the quality of the fit. 
They use Bayes' theorem to learn the distributions of both network parameters,
$\vec{\theta}    = \{ \mat{W}_k, \mat{b}_k \}$ and physics parameters, 
$\vec{\vartheta} = \{ c, k, x_0 \}$, 
with means $\mu$ and standard deviations $\sigma$ each, 
resulting in a total of 
2312 unknowns.
Figure \ref{fig10}, bottom right, suggests that 
Bayesian Physics Informed Neural Networks
provide a good fit of the training data and have a reasonable predictive potential outside the training window. The narrower credible intervals compared to the Bayesian Neural Network indicate that including physics increases the performance, especially in the prediction window. 
Bayesian Physics Informed Neural Networks are a powerful tool when used appropriately. They combine the advantages of all methods, and perform best, albeit at a larger computational cost. 
Since they have the largest set of parameters, their true performance is sensitive to an appropriate scaling and pre-conditioning and to the size of the training data set, as we can conclude from their convergence curve in Figure~\ref{fig11}. 
However, when scaled appropriately, in our example such that the parameter $k$ associated with the unknown $x(t)$ becomes unity, their performance increases drastically, as we conclude from the narrow credible intervals of the scaling study in Figure \ref{fig_conditioning}. 
\\[6.pt]
%%%%%%%%%%%%%%%%%%%%%%%%%%%%%%%%%%%%%%%%%%%%%%%%%%%%%%%%%%%%%%%%%%%
\noindent
Taken together, our study has shown that by embedding physical principles into a neural network architecture, we can generate effective models that train well, even with a small amount of data. 
Physics-Informed Neural Networks seamlessly integrate data and physics, robustly solve forward and inverse problems, and perform well for both interpolation and extrapolation. 
At only minor additional cost, they can self-adaptively learn the weighting between data and physics and smoothly integrate real-world data and physics-based modeling.
Combined with Bayesian Neural Networks, Physics-Informed Neural Networks can serve as a prior in a general Bayesian framework, and can generate estimators for posterior distributions that provide valuable insight into uncertainty quantification. 
Here we have only demonstrated these features for the simple model problem of a seasonal endemic infectious disease, but it is easy to see how the underlying concepts and trends would generalize to more complex disease conditions and, more broadly, to a wide variety of nonlinear dynamical systems.  
%%%%%%%%%%%%%%%%%%%%%%%%%%%%%%%%%%%%%%%%%%%%%%%%%%%%%%%%%%%%%%%%%%%
\section*{Acknowledgments}
\noindent
This work was supported 
by a DAAD Fellowship to Kevin Linka, 
by the  MURI/ARO grant W911NF-15-1-0562 to George Karniadakis, and
by the Stanford School of Engineering COVID-19 Research and Assistance Fund and Stanford Bio-X IIP seed grant to Ellen Kuhl.
%%%%%%%%%%%%%%%%%%%%%%%%%%%%%%%%%%%%%%%%%%%%%%%%%%%%%%%%%%%%%%%%%%%
%\bibliographystyle{elsarticle-num} 
%\bibliography{cas-refs}

\end{document}